\documentclass[research]{elsarticle}

\usepackage{hyperref}
\usepackage{amsmath}
\usepackage{amssymb}
\usepackage{multicol}
\usepackage{multirow}
\usepackage{bbding}
\usepackage{booktabs}
\usepackage{algorithm}
\usepackage{algorithmic}
\usepackage{graphicx}
\usepackage{epstopdf}
\usepackage{subfigure}
\usepackage{color}
\usepackage{xcolor}
\usepackage{enumitem}
\usepackage{bm}
\usepackage{enumitem}
\usepackage[perpage,symbol]{footmisc}
\setfnsymbol{wiley}
\journal{Journal of \LaTeX\ Templates}

\begin{document}

\begin{frontmatter}

\title{Cross-position Activity Recognition with Stratified Transfer Learning}

\author{Yiqiang Chen$^{ab}$\footnote{Corresponding author. Email address: {\ttfamily yqchen@ict.ac.cn}.}}
\author{Jindong Wang$^{ab}$}
\author{Meiyu Huang$^{c}$}
\author{Han Yu$^{d}$}

\address{$^a$Beijing Key Lab of Mobile Computing and Pervasive Devices, Institute of Computing Technology, Chinese Academy of Sciences}

\address{$^b$University of Chinese Academy of Sciences}
\address{$^c$Qian Xuesen Lab of Space Technology, China Academy of Space Technology}
\address{$^d$School of Computer Science and Engineering, Nanyang Technological University}

\begin{abstract}
Human activity recognition (HAR) aims to recognize the activities of daily living by utilizing the sensors attached to different body parts. The great success of HAR has been achieved by training machine learning models using sufficient labeled activity data. However, when the labeled data from a certain body position (i.e. \textit{target} domain) is missing, how to leverage the data from other positions (i.e. \textit{source} domain) to help recognize the activities of this position? This problem can be divided into two steps. Firstly, when there are several source domains available, it is often difficult to select the most \textit{similar} source domain to the target domain. Secondly, with the selected source domain, we need to perform accurate knowledge transfer between domains in order to recognize the activities on the target domain. Existing methods only learn the global distance between domains while ignoring the local property. In this paper, we propose a \textit{Stratified Transfer Learning} (STL) framework to perform both source domain selection and activity transfer. STL is based on our proposed \textit{Stratified} distance to capture the local property of domains. STL consists of two components: 1) Stratified Domain Selection (STL-SDS), which can select the most similar source domain to the target domain; and 2) Stratified Activity Transfer (STL-SAT), which is able to perform accurate knowledge transfer. Extensive experiments on three public activity recognition datasets demonstrate the superiority of STL. Furthermore, we extensively investigate the performance of transfer learning across different degrees of similarities and activity levels between domains. We also discuss the potential applications of STL in other fields of pervasive computing for future research.

\end{abstract}


\end{frontmatter}

\section{Introduction}
Human activity recognition~(HAR) aims to seek high-level knowledge from the low-level sensor inputs~\cite{sadri2017information}. For example, we can detect if a person is walking or running using the on-body sensors such as the smartphone or the wristband. HAR has been widely used to applications such as smart care~\cite{huber2017smartcare}, wireless sensing~\cite{asudeh2016general,shao2018traveling}, adaptive systems~\cite{krupitzer2015survey}, and smart home sensing~\cite{ogunnaike2017toward,wen2016adaptive}. 

Activities are of great importance to a person's health status. When a person is performing some activities, each of his body parts has certain activity patterns. Thus, sensors can be attached on some body positions to collect activity data which can be used to build machine learning models in order to recognize their activities. The combination of sensor signals from different body positions can be used to reflect meaningful knowledge such as a person's detailed health conditions~\cite{hammerla2015pd} and working states~\cite{plotz2011feature}. Unfortunately, it is nontrivial to designing wearing styles for a wearable device. On the one hand, it is not bearable to equip all the body positions with sensors which makes the activities not natural. Therefore, we can only attach sensors on limited body positions. On the other hand, it is impossible to perform HAR if the labels on some body positions are missing, since the activity patterns on specific body positions are significant to recognize certain information. 

\begin{figure}[htbp]
	\centering
	\includegraphics[scale=0.5]{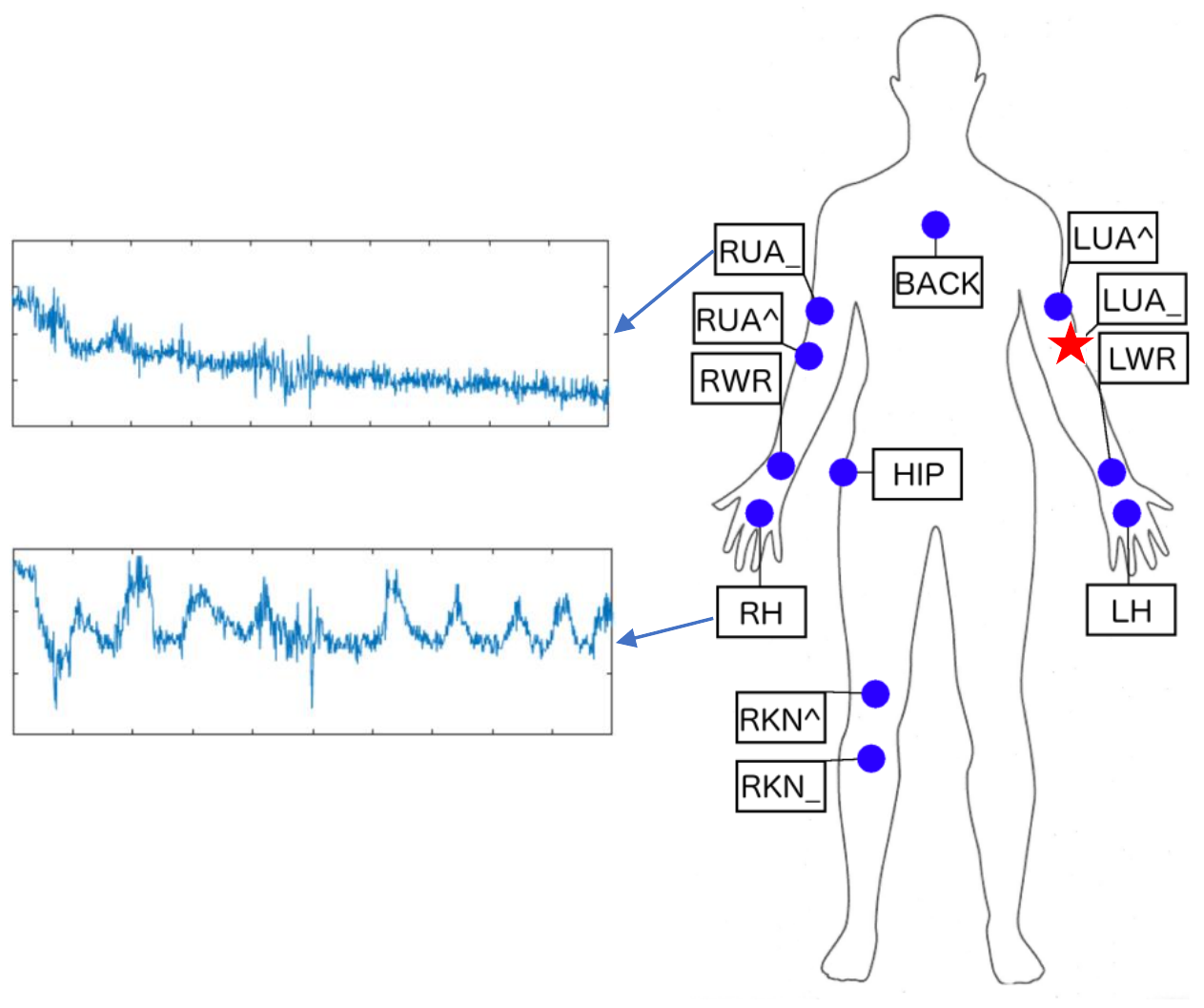}
	\caption{An example of cross-position activity recognition. Firstly, the activity signals on different body parts are often different. Secondly, if the labels of a certain part are missing (the red pentacle), how to leverage the well-labeled activity data on other body parts (the blue dots) to acquire its labels?}
	\label{fig-intro}
\end{figure}

Figure~\ref{fig-intro} illustrates this situation. Assume this person is suffering from Small Vessel Disease (SVD)~\cite{wardlaw2013neuroimaging}, which is a severe brain disease heavily related to activities. However, we cannot equip his all body with sensors to acquire the labels since this will make his activities unnatural. We can only label the activities on certain body parts in reality. If the doctor wants to see his activity information on the arm (the red pentacle, we call it the \textit{target domain}), which only contains sensor readings instead of labels, how to utilize the information on other parts (such as torso or leg, we call them the \textit{source domains}) to help obtain the labels on the target domain? This is referred to as the \textit{cross-position activity recognition (CPAR).} In this paper, we mainly focus on recognizing daily activities using sensors from a single body position.

The problem of CPAR is extremely challenging. Firstly, we do not know which body part is the most similar to the target position since the sensor signals are not independent, but highly correlated because of the shared body structures and functions. If we use all the body parts as the source domain, there is likely to be \textit{negative} transfer~\cite{pan2010survey} because some body parts may be dissimilar. Secondly, we only have the raw activity data on the target domain without the actual activity labels, making it infeasible to measure the similarities between different body positions. Thirdly, even when we know the similar body parts to the target domain, it is still difficult to build a good machine learning model using both the source and the target domains. The reason is that signals from different domains are following different distributions, which means there are distribution discrepancies between them. However, traditional machine learning models are built by assuming that all signals follow the same distribution. Fourthly, when it comes to \textit{multiple} persons, the sensor readings are more different compared to different body parts on one person. This makes the problem more challenging.

To tackle the above challenges, several transfer learning methods have been proposed~\cite{cook2013transfer}. The key is to learn and reduce the distribution divergence (distance) between two domains. With the distance, we can perform source domain selection as well as knowledge transfer. Based on this principle, existing methods can be summarized into two categories: exploiting the correlations between features~\cite{blitzer2006domain,kouw2016feature}, or transforming both the source and the target domains into a new shared feature space~ \cite{pan2011domain,gong2012geodesic,long2015domain}. 

Existing approaches tend to reduce the \textit{global} distance by projecting all samples in both domains into a single subspace. However, they fail to consider the \textit{local} property within classes~\cite{lin2016cross}. The \textit{global} distance may result in loss of domain local property such as the source label information and the similarities within the same class. Therefore, it will generate a negative impact on the source selection as well as the transfer learning process. It is necessary to exploit the local property of classes to overcome the limitation of global distance learning.

In this paper, we propose a \textbf{Stratified Transfer Learning} (STL) framework to tackle the challenges of both source domain selection and knowledge transfer in CPAR. The term \textit{stratified} comes from the notion of \textit{spliting at different levels and then combining}. We adopt the well-established assumption that data samples within the same class should lay on the same subspace, even if they come from different domains~\cite{elhamifar2013sparse}. Thus, \textit{stratified} refers to \textit{transformed subspace}. This has motivated us to propose the concept of \textbf{Stratified} distance (SD) in comparison to traditional \textit{Global} distance (GD). STL has two components regarding the challenges in CPAR: a \textbf{Stratified Domain Selection} (STL-SDS) algorithm to select the most similar source domain to the target domain, and a \textbf{Stratified Activity Transfer} (STL-SAT) method to perform activity knowledge transfer between different body parts. Both STL-SDS and STL-SAT are able to exploit the local property of domains, thus they can achieve promising results in CPAR. Comprehensive experiments on three large public activity recognition datasets (i.e. OPPORTUNITY, PAMAP2, and UCI DSADS) demonstrate that STL-SDS is better than the existing global distance approaches used in selecting source domains. STL-SAT outperforms five state-of-the-art methods with a significant improvement of \textbf{7.7}\% in classification accuracy with improved $F1$ score.

\textbf{Contributions.} Our contributions are four-fold:

1) We propose the \textit{Stratified Transfer Learning} (STL) framework for source domain selection and knowledge transfer in CPAR. STL is the \textit{first} attempt to exploit the \textit{Stratified} distance (SD) in order to capture the \textit{local} property between domains. SD is a general distance that can be applied to different transfer learning applications.

2) We propose the \textit{Stratified Domain Selection} (STL-SDS) algorithm to accurately select the most similar source domain to the target domain. Experiments demonstrate the superiority of STL-SDS compared to the traditional global distance measure.

3) We propose the \textit{Stratified Activity Transfer} (STL-SAT) method to perform knowledge transfer for cross-position activity recognition. Experiments demonstrate significantly improved accuracy achieved by STL-SAT compared to five state-of-the-art methods.

4) We extensively investigate the performance of cross-position activity recognition on different degrees of position similarity and different levels of activities. And we additionally discuss the potential of STL in other pervasive computing applications, providing experience for future research.

This paper is an extended version of our PerCom paper~\cite{wang2018stratified}, where we proposed a stratified transfer learning algorithm for activity transfer. That algorithm is regarded as STL-SAT in this paper. Beyond that, we further extend the idea of stratified learning and propose the stratified distance as the similarity measurement for domains. Based on this, we propose the STL-SDS algorithm for source domain selection. On the top, we unite STL-SDS and STL-SAT into a framework for CPAR and conduct extensive experiments to evaluate their performance.

The rest of this paper is organized as follows. Section~\ref{sec-related} reviews the related work. Section~\ref{sec-stl} introduces the proposed stratified transfer learning framework. In Section~\ref{sec-exp}, we present experimental evaluation and analysis on public datasets. In Section~\ref{sec-dis}, we discuss the potential of the framework in other real applications. Finally, the conclusions and future work are presented in Section~\ref{sec-conclu}.

\section{Related Work}
\label{sec-related}

\subsection{Activity Recognition}
Human Activity recognition has been a popular research topic in pervasive computing~\cite{bulling2014tutorial} for its competence in learning profound high-level knowledge about human activity from raw sensor inputs. Several survey articles have elaborated on the recent advance of activity recognition using conventional machine learning~\cite{bulling2014tutorial,lara2013survey} and deep learning~\cite{wang2017deep} approaches.

Conventional machine learning approaches have made tremendous progress on HAR by adopting machine learning algorithms such as similarity-based approach~\cite{zheng2011user,chen2016ocean}, active learning~\cite{hossain2016active}, crowdsourcing~\cite{lasecki2013real}, and other semi-supervised methods~\cite{nguyen2015did,hu2016less}. Those methods typically treat HAR as a standard time series classification problem. And they tend to solve it by subsequently performing preprocessing procedures, feature extraction, model building, and activity inference. However, they all assume that the training and test data are with the same distribution. As for CDAR where the training and the test data are from different feature distributions, those conventional methods are prune to under-fitting since their generalization ability will be undermined~\cite{pan2010survey}.

Deep learning based HAR~\cite{wang2017deep,wang2018deep} achieves the state-of-the-art performance than conventional machine learning approaches. The reason is that deep learning is capable of automatically extracting high-level features from the raw sensor readings~\cite{lecun2015deep}. Therefore, the features are likely to be more domain-invariant and tend to perform better for cross-domain tasks. A recent work evaluated deep models for cross-domain HAR~\cite{morales2016deep}, which provides experience on this area. In this paper, we mainly focus on the traditional approaches.

\subsection{Transfer Learning}
Transfer learning has been applied in many applications such as Wi-Fi localization~\cite{pan2008transfer}, natural language processing~\cite{blitzer2006domain}, and visual object recognition~\cite{duan2012domain}. According to the literature survey~\cite{pan2010survey}, transfer learning can be categorized into 3 types: instance-based, parameter-based, and feature-based methods. 

Instance-based methods perform knowledge transfer mainly through instance re-weighting techniques~\cite{tan2017distant,chattopadhyay2012multisource}. Parameter-based methods~\cite{yao2010boosting,zhao2011cross} first train a model using the labeled source domain, then perform clustering on the target domain.  Our framework belongs to the feature based category, which brings the features of source and target domain into the same subspace where the data distributions can be the same. A fruitful line of work has been done in this area~\cite{wang2017balanced,long2013transfer,long2015domain}. STL differs from existing feature-based methods in the following aspects:

\textbf{Exploit the correlations between features.} \cite{blitzer2006domain} proposed structural correspondence learning~(SCL) to generatively learn the relation of features. \cite{kouw2016feature} applied a feature-level transfer model to learn the dependence between domains, then trained a domain-adapted classifier. Instead of modeling the relationship of domain features, SAT transforms the domain data into a new subspace, which does not depend on the domain knowledge in modeling features.

\textbf{Transform domains into new feature space.} \cite{pan2008transfer} proposed maximum mean discrepancy embedding~(MMDE) to learn latent features in the reproducing kernel Hilbert space~(RKHS). MMDE requires solving a semidefinite programming~(SDP) problem, which is computationally prohibitive. \cite{pan2011domain} extended MMDE by Transfer Component Analysis~(TCA), which learns a kernel in RKHS. \cite{dorri2012adapting} adopted a similar idea. \cite{seah2012learning} learns target predictive function with a low variance. \cite{glorot2011domain} sampled the domain features by viewing the data in a Grassmann manifold to obtain subspaces. \cite{gong2012geodesic} exploited the low-dimensional structure to integrate the domains according to geodesic flow kernel~(GFK). Long \textit{et al.} proposed joint distribution adaptation~(JDA) based on minimizing joint distribution between domains, while SAT focuses on the marginal distribution. \cite{long2015domain} proposed transfer kernel learning~(TKL), which learned a domain-invariant kernel in RKHS. \cite{gong2016domain} studied the conditional transfer components between domains. Methods in these literature tend to learn some common representations in the new feature space, then a global domain shift can be achieved. However, the difference between individual classes is ignored.

\subsubsection{Source Domain Selection}

The work~\cite{xiang2011source} first proposed a source-selection free transfer learning approach. They choose the source samples that are close to the target samples using the Laplacian Eigenmap. The work \cite{lu2014source} followed this idea in the text classification. However, both of them only focused on the sample selection, while our STL-SDS focuses on the selection of the whole domain. Collier \textit{et al.}~\cite{collier2018cactusnets} investigated the transfer performance of different layers of a neural network in a grid search manner. But they did not perform source selection. The work~\cite{sung2017learning} developed a relation network, which can be used to evaluate the distance between different image samples. Yet they still focused on the global distance. Authors in \cite{bhatt2016multi} proposed a greedy multi-source selection algorithm. This selection algorithm could iteratively select the best $K$ source domains and then perform transfer learning based on this selection. However, their method still relies on the similarity calculated by the global distance. Our STL-SDS is the first work to perform source domain selection using the local property.

\subsubsection{Transfer Learning based Activity Recognition}
Some existing work also focused on transfer learning based HAR~\cite{cook2013transfer}. Among existing work, Zhao \textit{et al.} proposed a transfer learning method TransEMDT~\cite{zhao2011cross} using decision trees, but it ignored the intra-class similarity within classes. \cite{khan2017transact} proposed the TransAct framework, which is a boosting-based method and ignores the feature transformation procedure. Thus it is not feasible in most activity cases. Feuz \textit{et al.}~\cite{feuz2017collegial} proposed a heterogeneous transfer learning method for HAR, but it only learns a global domain shift. A more recent work of Wang \textit{et al.}~\cite{wang2018deep} performs activity recognition using deep transfer learning, which also focused on the global distance between domains. To the best of our knowledge, our STL framework is the first attempt towards learning a stratified distance to capture the local property of domains.

\section{Stratified Transfer Learning}
\label{sec-stl}

In this section, we introduce the Stratified Transfer Learning~(STL) framework for cross-position activity recognition.

\subsection{Problem Definition}
\label{sec-problem}

The goal of cross-position activity recognition~(CPAR) is to predict the activities of one body part (i.e. the \textit{target} domain) using existing labeled data from another body part (i.e. the \textit{source} domain). Formally speaking, we can use $\mathcal{D}_{s} = \{(\mathbf{x}^s_{i},y^s_{i})\}^{n_s}_{i=1}$ to denote the labeled source domain, and $\mathcal{D}_{t} = \{(\mathbf{x}^t_{j},y^t_j)\}^{n_t}_{j=1}$ denotes the target domain. Note that $y^t_j$ is missing, i.e. it is the goal of CPAR. In CPAR, we assume the sensor modalities and the activity categories are equal in all body parts, thus the feature space $\mathcal{X}^s=\mathcal{X}^t$ and label space $\mathcal{Y}_s = \mathcal{Y}_t$. Here, $d$ denotes the feature dimension. We use $C$ to denote the total categories of both domains.  The different data distributions on different body parts means that the marginal and conditional distributions between domains are different. Therefore, $P(\mathbf{x}_{s}) \neq P(\mathbf{x}_{t})$, and $Q(y_{s} | \mathbf{x}_{s}) \neq Q(y_{t}|\mathbf{x}_t)$.

Note that we only focus on the setting where the source and the target domains are sharing the same label spaces (i.e. $\mathcal{Y}_s = \mathcal{Y}_t$) since this setting is rather common in transfer learning research~\cite{pan2010survey} and many previous work~\cite{zhao2010cross,zhao2011cross}. This is likely to be a class imbalance problem in theory when some particular classes are with extremely few samples. However, since our setting is based on the common activities of daily living where there exist enough samples for each class, we will leave class imbalance problem as a future work.

This problem is extremely challenging. We neither know which body part is the most similar to the target domain, nor we know how to build a machine learning model to deal with the different distributions. Accordingly, we separate it into two steps: 1) Determine the most similar source domain to the target domain. 2) Transfer the knowledge from the source domain to the target domain. 

\subsection{Motivation}
The key to successful activity transfer learning is to utilize the \textit{similarity} between the source and the target domain~\cite{pan2010survey}. It is not difficult to directly measure the similarity between domains using existing distance functions. For instance, we can adopt the Euclidean distance to compute the sample-wise distance and then average them; or we can use the Kullback–-Leibler divergence~\cite{wang2017balanced} to calculate the probability distance between domains. Unfortunately, these computing approaches are only calculating the \textit{global} distance between domains since they are applied to the whole domain. The \textit{global} distance may result in loss of domain local property such as the source label information and the similarities within the same class. Therefore, it will generate a negative impact on the source selection as well as the transfer learning process.

In order to capture the local property of domains, we adopt the assumption~\cite{elhamifar2013sparse}: the data samples from the same class should lay on the \textit{same} subspace, even if they belong to different domains. By following this assumption, we propose the \textit{stratified} distance to capture the local property of domains. The \textit{stratified} distance refers to the class-wise distance between two different domains. For example, if there are 5 classes in two domains: $\mathcal{D}_s = (\mathcal{D}^{(1)}_s,\ldots,\mathcal{D}^{(5)}_s)$ and $\mathcal{D}_t=(\mathcal{D}^{(1)}_t,\ldots,\mathcal{D}^{(5)}_t)$, then the stratified distance (SD) should be calculated as:
\begin{equation}
\label{eq-local}
	SD = \frac{1}{5} \sum_{i=1}^{5} Dist(\mathcal{D}^{(i)}_s, \mathcal{D}^{(i)}_t),
\end{equation}
where $Dist(\cdot)$ denotes some distance function such as the Euclidean distance. As a comparison, the global distance (GD) can be represented as:
\begin{equation}
\label{eq-global}
	GD = Dist(\mathcal{D}_s,\mathcal{D}_t).
\end{equation}

\begin{figure}[t!]
	\centering
	\subfigure[Original target]{
		\centering
		\includegraphics[scale=0.42]{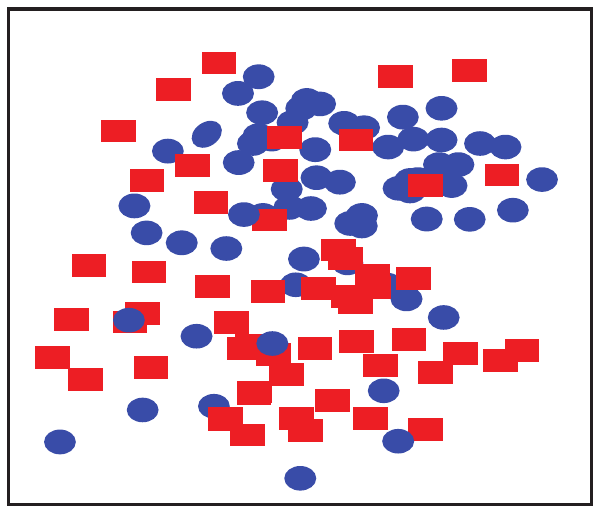}
		\label{fig-sub-target}}
	\hspace{.2in}
	\subfigure[Global dist.]{
		\centering
		\includegraphics[scale=0.42]{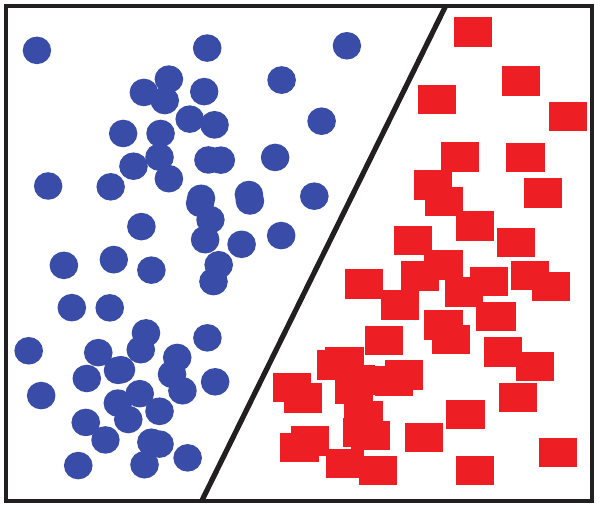}
		\label{fig-sub-global}}
	\hspace{.2in}
	\subfigure[Stratified dist.]{
		\centering
		\includegraphics[scale=0.42]{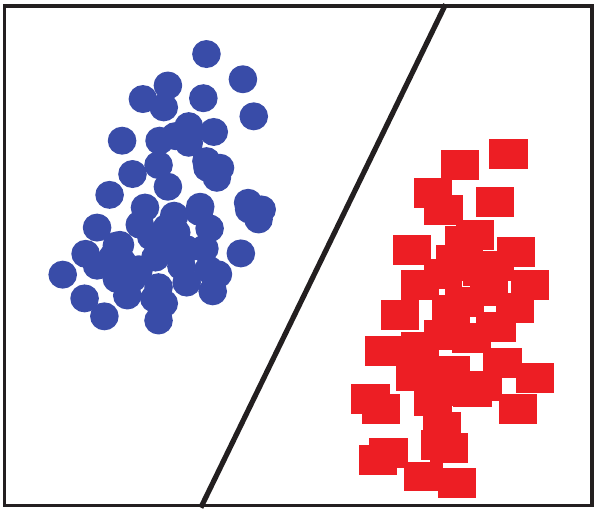}
		\label{fig-sub-stl}}
	
	\caption{Traditional global distance (GD) and proposed stratified distance (SD).}
	\label{fig-motiv}
\end{figure}

Figure~\ref{fig-motiv} briefly illustrates the results of \textit{global} and our \textit{stratified} distance using two classes of the same target domain~\cite{wang2018stratified}. It indicates that $SD$ could not only help to learn good classification function, but helps to obtain tighter within-class distance. 

\subsection{The STL Framework}
We can exploit the proposed Stratified distance (SD) measure to perform source domain selection and activity transfer in CPAR. Technically, there are two critical challenges ahead. Firstly, SD is based on the labels of both domains, while the target domain has no labels. Secondly, even if we could obtain the labels for the target domain, it is nontrivial to solve Eq.~\ref{eq-intrammd} since we do not know how to choose the $Dist(\cdot)$ function.

\begin{figure}[htbp]
	\centering
	\setlength{\fboxrule}{1pt} 
	\setlength{\fboxsep}{0.1cm}
	\fbox{\includegraphics[scale=0.45]{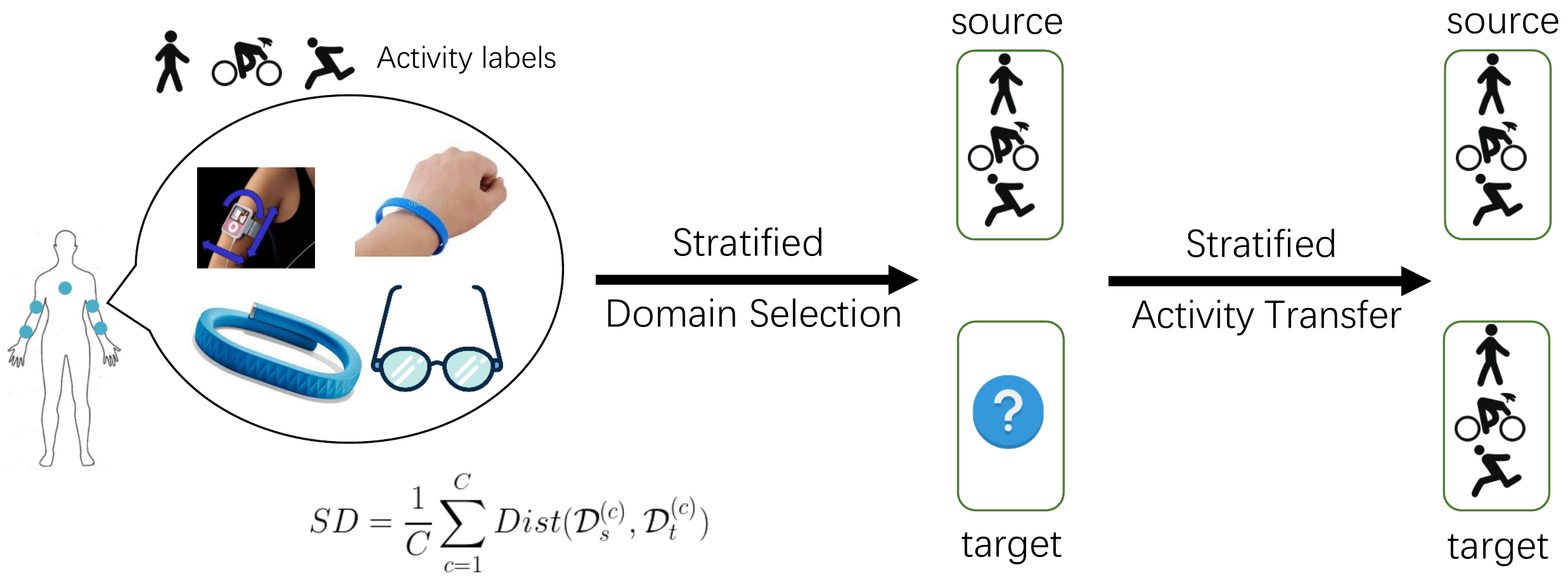}}
	\caption{Main idea of the Stratified Transfer Learning~(STL) framework. There are two steps: (1)~Stratified Domain Selection (STL-SDS), which can select the most similar source domains to the target domains. (2)~Stratified Activity Transfer (STL-SAT), which can perform activity recognition on the target domain based on transfer learning. Both STL-SDS and STL-SAT are well exploiting the idea of our proposed stratified distance.}
	\label{fig-stl-framework}
\end{figure}

In this paper, we propose the \textbf{Stratified Transfer Learning} (STL) framework to address both of these two challenges. Firstly, STL uses a popular \textit{majority voting} technique to acquire the labels for the target domain. Secondly, STL exploits an effective distance measure called Maximum Mean Discrepancy (MMD) to calculate the SD distance and perform \textit{Intra-class transfer}. Finally, STL could perform source domain selection and activity transfer based on the calculated distance. Figure~\ref{fig-stl-framework} illustrates the main idea of the STL framework.

\subsubsection{Majority voting}

We can easily obtain the \textit{pseudo} labels for the target domain based on a weak classier such as 1-nearest-neighbor (1NN) trained on the source domain by following existing methods~\cite{long2013transfer}. However, we consider that since there is a large distribution discrepancy between two domains, a simple 1NN classifier may not work well. Therefore, we propose to use the \textit{majority voting} technique to exploit the knowledge from the crowd~\cite{prelec2017solution}. The idea is that one certain classifier may be less reliable, so we assemble several different classifiers to obtain more reliable pseudo labels. To this end, STL makes use of some base classifiers learned on $\mathcal{D}_s$ to collaboratively learn the labels for $\mathcal{D}_t$.

Let $A_j (j = 1,2,\cdots,n_2)$ denotes the final result of majority voting on $\mathbf{x}_{t_j}$, and $f_t(j)$ denotes the prediction of the $j$-th sample by the $t$-th classifier $f_t(\cdot)$, then:
\begin{equation}
\label{equ-majority}
A_j = \begin{cases}
\mathrm{majority}(f_t(j), t) &  \text{if majority holds}\\
-1 & \text{otherwise}
\end{cases}
\end{equation}
where $t \in \{1,2,\cdots\}$ denotes the index of the classifier.

The majority voting technique generally ensembles all the classifiers learned in the source domain. The condition ``majority holds'' refers to any potential scheme that helps to generate a better solution such as simple voting and weighted voting. Specifically, $A_j$ could be defined as a) if most classifiers have the same results on a sample, we take its label, else we label it `-1'; b) same as a) with voting weights to classifiers; c) the stacking of some base classifiers. In theory, the classifiers can be of any type.

Formally, we call the samples with pseudo labels in the target domain \textit{Candidates}, which is denoted as $\mathbf{X}_{can}$. For those samples with label `-1', we call them \textit{Residuals} and denote as $\mathbf{X}_{res}$.

Using majority voting, STL can generate reliable \textit{pseudo} labels for the target domain.  These pseudo labels will act as evidence of the following transfer learning algorithms. The effectiveness of this technique will be evaluated in later sections.

\subsubsection{Intra-class transfer}

In this step, STL exploits the local property of domains to further transform each class of the source and target domains into the same subspace. Since the properties within each class are more similar, the \textit{Intra-class transfer} technique will guarantee that the transformed domains have the minimal distance. In this step, we only consider the \textit{candidates} from the target domain for their reliable pseudo labels.

Initially, $\mathcal{D}_s$ and $\mathbf{X}_{can}$ are divided into $C$ groups according to their~(pseudo) labels, where $C$ is the total number of classes. Then, feature transformation is performed within each class of both domains. Finally, the results of distinct subspaces are merged.

In order to achieve intra-class transfer, we need to calculate the distance between \textit{each class}. Since the target domain has no labels, we use the pseudo labels from majority voting. For the candidates and the source domain, we calculate their \textit{intra-class} distance using the \textit{stratified} distance:
\begin{equation}
\label{eq-intrammd}
SD = \frac{1}{C} \sum_{c=1}^{C} Dist(\mathcal{D}^{(c)}_s, \mathbf{X}^{(c)}_{can}),
\end{equation}
where $\mathcal{D}^{(c)}_s$ and $\mathbf{X}^{(c)}_{can}$ denote the samples from class $c$ in the source and candidates, respectively.

This distance can easily be calculated using existing metrics such as Euclidean distance and Kullback-Leibler (KL) divergence~\cite{wang2017balanced}. However, the Euclidean distance and KL divergence are too general for activity recognition problem, which ignores the label information of the source domain. Moreover, the KL divergence needs to first estimate the probability of both domains, which is trivial and not suitable since the target domain has no labels. Therefore, we need to calculate the similarity between domains while considering the label information on the source domain.

In order to calculate the function $Dist(\cdot)$ in Eq.~\ref{eq-local}, we adopt Maximum Mean Discrepancy~(MMD)~\cite{gretton2012kernel} as the measurement. MMD is a nonparametric method to measure the divergence between two distinct distributions and it has been widely applied to many transfer learning methods~\cite{pan2011domain,long2015domain}. The MMD distance between two domains can be formally computed as:
\begin{equation}
\label{equ-mmd}
\begin{split}
D(\mathcal{D}_{s},\mathbf{X}_{can})
=\sum_{c=1}^{C}\left \Vert \frac{1}{n^{(c)}_s} \sum_{\mathbf{x}^{s}_{i} \in \mathcal{D}^{(c)}_s} \phi(\mathbf{x}^{s}_{i}) - \frac{1}{n^{(c)}_t} \sum_{\mathbf{x}^{t}_{j} \in \mathbf{X}^{(c)}_{can}} \phi(\mathbf{x}^{t}_{j}) \right \Vert ^2_\mathcal{H}
\end{split},
\end{equation}
where $\mathcal{H}$ denotes reproducing kernel Hilbert space~(RKHS). $n^{(c)}_s=|\mathcal{D}^{(c)}_s|, n^{(c)}_t=|\mathbf{X}^{(c)}_{can}|$. Here $\phi(\cdot)$ denotes some feature map to map the original samples to RKHS. The reason we do not use the original data is that the features are often distorted in the original feature space, and it can be more efficient to perform knowledge transfer in RKHS~\cite{pan2011domain}. 

Therefore, given a predefined mapping function $\phi(\cdot)$, we can compute the intra-class distance between two domains.

\textbf{Remark:} The majority voting and intra-class transfer are common steps in STL. In the next sections, we will elaborate on how to use the STL framework to perform source domain selection and activity transfer.

\subsection{Stratified Domain Selection}

Given a set of body parts with label information, we have to determine the body part that has the most similar property to the target body part. In this paper, based on the STL framework, we propose the \textit{Stratified Domain Selection (STL-SDS)} algorithm to handle this challenge. Based on the proposed \textit{Stratified} distance, STL-SDS well exploits the local property of different domains as well as the supervised information on the source domain. 

We adopt a \textit{greedy} technique in STL-SDS. We know that the most similar body part to the target is the one with the most similar structure and body functions. Therefore, we use the distance to reflect their similarity. We calculate the stratified distance according to Eq.~\ref{equ-mmd} between each source domain and the target domain and select the one with the minimal distance.

Unfortunately, it is non-trivial to solve Eq.~\ref{equ-mmd} directly since the mapping function $\phi(\cdot)$ is to be determined. Simply use a certain function will ruin the local property of domains. Thus, we turn to some kernel methods. We define a kernel matrix $\mathbf{K} \in \mathbb{R}^{(n_1+n_2)\times(n_1+n_2)}$, which can be constructed by the inner product of the mapping:
\begin{equation}
\label{equ-kernel}
\mathbf{K}_{ij}=\langle\phi(\mathbf{x}^{s}_i),\phi(\mathbf{x}^{t}_j)\rangle = \phi(\mathbf{x}^{s}_i)^\top \phi(\mathbf{x}^{t}_j),
\end{equation}
where $\mathbf{x}^{s}_i$ and $\mathbf{x}^{t}_j$ are samples from either $\mathcal{D}_s$ or $\mathbf{X}_{can}$.

Therefore, we can easily calculate this equation without losing the local property by giving $\phi(\cdot)$ a certain implementation. In our work, we use the Radical Basis Funcation (RBF), which can be defined as:

\begin{equation}
	\label{eq-rbf}
	K(\mathbf{x}^{s}_i,\mathbf{x}^{t}_j) = \exp \left(- \frac{||\mathbf{x}^{s}_i - \mathbf{x}^{t}_j||^2_2}{2 \sigma^2}\right),
\end{equation}
where $\sigma$ is the kernel bandwidth.

The complete learning process of STL-SDS is in Algorithm~\ref{algo-sds}.

\begin{algorithm}[htbp]
	\caption{STL-SDS:~\underline{S}tratified \underline{D}omain \underline{S}election}
	\label{algo-sds}
	\renewcommand{\algorithmicrequire}{\textbf{Input:}} 
	\renewcommand{\algorithmicensure}{\textbf{Output:}} 
	\begin{algorithmic}[1]
		\REQUIRE~
		A list of source domains $\mathcal{D}_{s_1},\cdots,\mathcal{D}_{s_K}$, target domain $\mathcal{D}_{t}$.\\
		\ENSURE~
		The source domain that has the minimal distance to $\mathcal{D}_t$.\\
		\STATE Initialize a source domain set $S=\{\}$;
		\STATE Perform majority voting on $\mathcal{D}_t$ to acquire $\mathbf{X}_{can}$;
		\STATE Obtain the pseudo labels $\mathbf{y}_t$ for $\mathcal{D}_t$;
		\FOR{i = 1 to K}
		\STATE Compute the distance $SD_i$ between $\mathcal{D}_{s_i}$ and $\mathbf{X}_{can}$ using Eq.~\ref{equ-mmd};
		\STATE Add $SD_i$ to S;
		\ENDFOR
		\RETURN Index $j$ of source domain that $SD_j=\min \{S\}$.
	\end{algorithmic}
\end{algorithm}

\subsection{Stratified Activity Transfer}
\begin{figure}[htbp]
	\centering
	\setlength{\fboxrule}{1pt} 
	\setlength{\fboxsep}{0.1cm}
	\fbox{
		\includegraphics[scale=0.68]{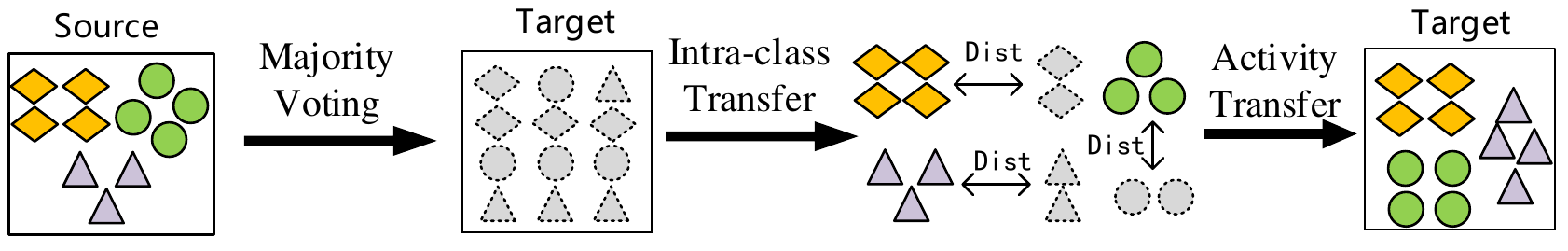}}
	\caption{Main idea of Stratified Activity Transfer (STL-SAT). There are three steps: (1)~Candidates generating to generate pseudo labels for the target domain; (2)~Perform intra-class transfer between source domain and candidates; (3)~Perform activity transfer using the transferred data.}
	\label{fig-stl-sat}
\end{figure}

After source domain selection, we can obtain the most similar body part to the target domain. The next step is to design an accurate transfer learning algorithm to perform activity transfer. In this paper, we propose a Stratified Activity Transfer (STL-SAT) method for activity recognition. STL-SAT is also based on our \textit{stratified} distance, and it can simultaneously transform the individual classes of the source and target domains into the same subspaces by exploiting the local property of domains. After feature learning, STL can learn the labels for the candidates. Finally, STL-SAT will perform a second annotation to obtain the labels for the residuals. The process of STL-SAT can be seen in Figure~\ref{fig-stl-sat}.

The feature transformation is also based on the intra-class transfer technique in Eq. \ref{equ-mmd}. However, different from STL-SDS where we use a certain mapping function such as RBF for feature mapping, we \textit{learn} this mapping $\phi(\cdot)$ in this step. The reason is that STL-SDS only needs one particular feature map to avoid the feature distortion, while STL-SAT will find the \textit{optimal} feature map with the minimal domain distance.

The learning of this feature map also starts with Eq.~\ref{equ-mmd}. For efficient learning, we introduce a feature transformation matrix $\mathbf{W} \in \mathbb{R}^{(n_1+n_2)\times m}$ to transform the samples of both domains from the original space to the RKHS. Here $m \ll d$ denotes the dimension after feature transformation. Thus, learning $\phi(\cdot)$ is equal to learning the matrix $\mathbf{W}$. Then, by applying kernel tricks, Eq.~\ref{equ-mmd} can be eventually formulated as the following trace optimization problem:
\begin{equation}
\label{equ-stratified}
\begin{split}
\min_{\mathbf{W}} \quad &\sum_{c=1}^{C} \mathrm{tr} (\mathbf{W}^\top \mathbf{K} \mathbf{L}_c \mathbf{K} \mathbf{W}) + \lambda \mathrm{tr}(\mathbf{W}^\top \mathbf{W})\\
\text{s.t.} \quad & \mathbf{W}^\top \mathbf{K} \mathbf{H} \mathbf{K} \mathbf{W} = \mathbf{I}
\end{split}
\end{equation}

There are two terms in the objective function of Eq.~\ref{equ-stratified}. The first term ($\sum_{c=1}^{C} \mathrm{tr} (\mathbf{W}^\top \mathbf{K} \mathbf{L}_c \mathbf{K} \mathbf{W})$) denotes the MMD distance of each class between source and target domain, and the second one ($\lambda \mathrm{tr}(\mathbf{W}^\top \mathbf{W})$) denotes the regularization term to ensure the problem is well-defined with $\lambda$ the trade-off parameter. The constraint in Eq.~\ref{equ-stratified} is used to guarantee that the transformed data ($\mathbf{W}^\top \mathbf{K}$) will still preserve some structure property of the original data. $\mathbf{I}_{n_s+n_t}$ is the identical matrix, and $\mathbf{H} = \mathbf{I}_{n_s + n_t} - 1/(n_s + n_t)\mathbf{11}^\top$ is the centering matrix. For notational brevity, we will drop the subscript for $\mathbf{I}_{n_s+n_t}$ in the sequel. $\mathbf{L}_c$ is the intra-class MMD matrix, which can be constructed as:
\begin{equation}
\label{equ-lc}
(\mathbf{L}_c)_{ij}=\begin{cases}
\frac{1}{{(n^{(c)}_s)}^2} & \mathbf{x}_i,\mathbf{x}_j \in \mathcal{D}^{(c)}_s\\ 
\frac{1}{{(n^{(c)}_t)}^2} & \mathbf{x}_i,\mathbf{x}_j \in \mathbf{X}^{(c)}_{can}\\ 
-\frac{1}{n^{(c)}_s n^{(c)}_t} & \begin{cases}
\mathbf{x}_i \in \mathcal{D}^{(c)}_s, \mathbf{x}_j \in \mathbf{X}^{(c)}_{can}\\
\mathbf{x}_i \in \mathbf{X}^{(c)}_{can}, \mathbf{x}_j \in \mathcal{D}^{(c)}_s
\end{cases}\\
0 & \text{otherwise}
\end{cases}
\end{equation}

\textbf{Learning algorithm}: Acquiring the solution of Eq.~\ref{equ-stratified} is non-trivial. To this end, we adopt Lagrange method as most of the existing work did~\cite{pan2011domain,long2015domain}. We denote $\bm{\Phi}$ the Lagrange multiplier, then the Lagrange function can be derived as
\begin{equation}
\label{equ-lagrange}
\begin{split}
L = ~& \mathrm{tr} \left(\mathbf{W}^\top \mathbf{K} \sum_{c=1}^{C} \mathbf{L}_c \mathbf{K}^\top \mathbf{W} \right) + \lambda \mathrm{tr}(\mathbf{W}^\top \mathbf{W})\\
& + \mathrm{tr} \left((\mathbf{I}-\mathbf{W}^\top \mathbf{K} \mathbf{H} \mathbf{K}^\top \mathbf{W})\bm{\Phi} \right)
\end{split}
\end{equation}

Setting the derivative $\partial L/\partial \mathbf{W}=0$, Eq.~\ref{equ-lagrange} can be finally formalized as an generalized eigen-decomposition problem
\begin{equation}
\label{equ-eigen}
\left(\mathbf{K} \sum_{c=1}^{C} \mathbf{L}_c \mathbf{K}^\top + \lambda \mathbf{I} \right) \mathbf{W}
=\mathbf{K} \mathbf{H} \mathbf{K}^\top \mathbf{W} \bm{\Phi}
\end{equation}

\begin{algorithm}[t!]
	\caption{STL-SAT:~\underline{S}tratified \underline{A}ctivity \underline{T}ransfer}
	\label{algo-stl}
	\renewcommand{\algorithmicrequire}{\textbf{Input:}} 
	\renewcommand{\algorithmicensure}{\textbf{Output:}} 
	\begin{algorithmic}[2]
		\REQUIRE~
		Source domain $\mathcal{D}_{s}$, target domain $\mathcal{D}_{t}$, dimension $m$.\\
		\ENSURE~
		The labels for the target domain: $\{\mathbf{y}_{t}\}$.\\
		\STATE Perform majority voting on $\mathcal{D}_{t}$ using Eq.~\ref{equ-majority} to get $\{\mathbf{X}_{can},\widetilde{\mathbf{y}}_{can}\}$ and $\mathbf{X}_{res}$;
		\STATE Construct kernel matrix $\mathbf{K}$ according to Eq.~\ref{equ-kernel} using $\mathbf{X}_{src}$ and $\mathbf{X}_{can}$, and compute the intra-class MMD matrix $\mathbf{L}_c$ using Eq.~\ref{equ-lc};
		\REPEAT
		\STATE Solve the eigen-decomposition problem in Eq.~\ref{equ-eigen} and take the $m$ smallest eigen-vectors to obtain the transformation matrix $\mathbf{W}$;
		\STATE Transform the same classes of $\mathbf{X}_s$ and $\mathbf{X}_{can}$ into the same subspaces using $\mathbf{W}$, and then merge them;
		\STATE Perform second annotation to get $\{\hat{\mathbf{y}}_{can}\}$ and $\{\hat{\mathbf{y}}_{res}\}$;
		\STATE Construct kernel matrix $\mathbf{K}$ and compute the intra-class MMD matrix $\mathbf{L}_c$ using Eq.~\ref{equ-lc};
		\UNTIL{Convergence}
		\RETURN $\{\mathbf{y}_{t}\}$.
	\end{algorithmic}
\end{algorithm}

Solving Eq.~\ref{equ-eigen} refers to solve this generalized eigen-decomposition problem and take the $m$ smallest eigenvectors to construct $\mathbf{W}$. $\mathbf{W}$ can transform both domains into the same subspace with minimum domain distance while preserving their properties. Since the knowledge transfer pertains to each class, we call this step \textit{intra-class transfer}, and that is where the name~\textit{stratified} originates from. After this step, the source and target domains belonging to the same class are simultaneously transformed into the same subspaces. 

After solving the above optimization problem, it is easy to get more reliable predictions~($\hat{\mathbf{y}}_{can}$) of the \textit{candidates}. Specifically, we train a standard classifier using $\{[\mathbf{W}^\top \mathbf{K}]_{1:n_1,:},\mathbf{y}_{s}\}$ and apply prediction on $[\mathbf{W}^\top \mathbf{K}]_{n_1+1:n_2,:}$. Finally, the labels of \textit{residuals} can be obtained by training classifier on instances $\{\mathbf{X}_{can},\hat{\textbf{y}}_{can}\}$. The labels of candidates can be correspondingly close to the ground truth by annotating twice since the domains are now in the same subspace after intra-class transfer.

The overall process of STL-SAT is described in Algorithm~\ref{algo-stl}. 

Remark: It should be noted that STL-SAT could achieve a better prediction if we use the result of \textit{second annotation} as the initial state and run \textit{intra-class transfer} iteratively. This \textit{EM-like} algorithm is empirically effective and will be validated in the following experiments. Additionally, we \textit{only} use majority voting in the first round of the iteration, then the proposed SAT could iteratively refine the labels for the target domain by \textit{only} using its previous results.

\section{Experimental Evaluation}
\label{sec-exp}
In this section, we evaluate the performances of STL framework (STL-SDS and STL-SAT) via extensive experiments on public activity recognition datasets.

\subsection{Datasets and Preprocessing}
Three large public datasets are adopted in experiments. Table~\ref{tb-dataset} provides a brief introduction to them. In the following, we briefly introduce those datasets, and more information can be found in their original papers. OPPORTUNITY dataset (\textbf{OPP})~\cite{chavarriaga2013opportunity} is composed of 4 subjects executing different levels of activities with sensors tied to more than 5 body parts. PAMAP2 dataset (\textbf{PAMAP})~\cite{reiss2012introducing} is collected by 9 subjects performing 18 activities with sensors on 3 body parts. UCI daily and sports dataset (\textbf{DSADS})~\cite{barshan2014recognizing} consists of 19 activities collected from 8 subjects wearing body-worn sensors on 5 body parts. Accelerometer, gyroscope, and magnetometer are all used in these datasets.

\begin{table*}[htbp]
	\centering
	\vspace{-.1in}
	\caption{Statistical information of three public datasets for activity recognition}
	\label{tb-dataset}
	\resizebox{\textwidth}{!}{
		\begin{tabular}{|c|c|c|c|c|}
			\hline
			\textbf{Dataset} & \textbf{Person} & \textbf{Activity} & \textbf{Sample} & \textbf{Position} \\ \hline \hline
			OPP & 4 & 4 & 701K  & Back (B), Right Upper Arm (RUA), Right Left Arm (RLA), Left Upper Arm (LUA), Left Lower Arm (LLA) \\ \hline
			PAMAP & 9 & 18 & 284M  & Hand (H), Chest(C), Ankle (A) \\ \hline
			DSADS & 8 & 19 & 114M  & Torso (T), Right Arm (RA), Left Arm (LA), Right Leg (RL), Left Leg (LL) \\ \hline
		\end{tabular}
	}
	\vspace{-.1in}
\end{table*}

\begin{figure}[htbp]
	\centering
	\includegraphics[scale=0.48]{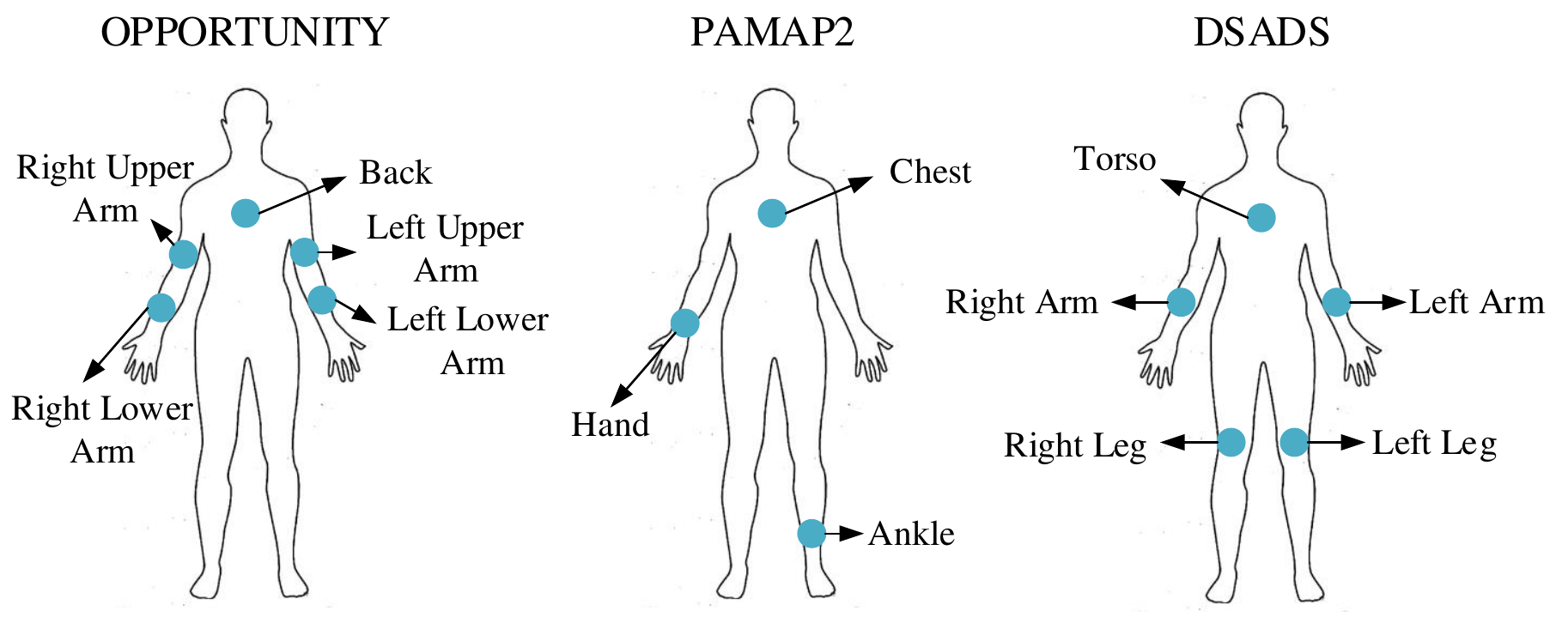}
	\caption{Different positions on OPPOPTUNITY, PAMAP2 and DSADS.}
	\label{fig-dataset}
\end{figure}

Figure~\ref{fig-dataset} illustrates the positions we investigated in three datasets. In our experiments, we use the data from all three sensors in each body part since most information can be retained in this way. For one sensor, we combine the data from 3 axes together using $a=\sqrt{x^2+y^2+z^2}$. Then, we exploit the sliding window technique to extract features~(window length is 5s). The feature extraction procedure is mainly executed according to existing work~\cite{hu2017okrelm}. In total, 27 features from both time and frequency domains are extracted for a single sensor. Since there are three sensors~(i.e. accelerometer, gyroscope, and magnetometer) on one body part, we extracted 81 features from one position. Table~\ref{tb-feature} shows the features we extracted for each dataset.

In order to better exploit these three datasets, we perform two aspects of cross-position activity recognition: \textit{Within Dataset} and \textit{Cross dataset}. \textit{Within Dataset} refers to perform CPAR inside a particular dataset. For instance, we can learn the labels for the Right Arm (RA) of the DSADS dataset by using labeled data from the other four body parts. In contrast, \textit{Cross dataset} uses the labeled body parts from another different dataset as the source domains. In our experiments, we use DSADS and OPP datasets for \textit{Within dataset} since they contain more body parts than PAMAP. We use all classes for each dataset in this aspect. For \textit{Cross Dataset}, we use the body parts from DSADS and OPP as the target domain, and use body parts from PAMAP as the source domain since there are more samples in PAMAP than other two datasets (Table~\ref{tb-dataset}). Note that there are different activities in three datasets. Thus we extract 4 common classes: \textit{Walking, Sitting, Lying}, and \textit{Standing}. 

\begin{table}[htbp]
	\centering
	\caption{Features extracted per sensor on each body part}
	\label{tb-feature}
		\begin{tabular}{@{}lll@{}}
			\toprule
			\textbf{ID}    & \textbf{Feature}                 & \textbf{Description}                                  \\ \midrule
			1     & Mean                    & Average value of samples in window           \\ 
			2     & STD                     & Standard deviation                           \\ 
			3     & Minimum                 & Minimum                                      \\ 
			4     & Maximum                 & Maximum                                      \\ 
			5     & Mode                    & The value with the largest frequency         \\
			6     & Range                   & Maximum minus minimum                        \\ 
			7     & Mean crossing rate      & Rate of times signal crossing mean value\\ 
			8     & DC                      & Direct component                             \\ 
			9-13  & Spectrum peak position  & First 5 peaks after FFT                      \\ 
			14-18 & Frequency               & Frequencies corresponding to 5 peaks\\ 
			19    & Energy                  & Square of norm                               \\ 
			20-23 & Four shape features     & Mean, STD, skewness, kurtosis                \\ 
			24-27 & Four amplitude features & Mean, STD, skewness, kurtosis                \\ \bottomrule
		\end{tabular}
\end{table}

\subsection{Evaluation of Stratified Domain Selection}
We perform source domain selection using the proposed STL-SDS (i.e. the stratified distance (SD)) algorithm on both the \textit{Within Dataset} and \textit{Cross Dataset} aspects. The comparison method is the \textit{global} distance (GD) using the MMD. 

It is worth noting that there is no effective evaluation metric for this domain selection problem. We can \textit{never} quantitatively know the actual distance between two activity domains. Thus, we evaluate the similarity based on \textit{classification accuracy} using the same classifier following existing work~\cite{wang2018deep}. For instance, for source domains \textit{A} and \textit{B} and target domain \textit{C}, we respectively train a linear SVM classifier using \textit{A} or \textit{B} and apply prediction on \textit{C}. The accuracy acts as the ground \textit{truth} for domain similarity: higher accuracy means shorter distance. Then, the source domain with the highest accuracy is the \textit{right} source for the target domain. Therefore, we can obtain the `ground truth' for the source domain selection.

\begin{figure}[htbp]
	\centering
	\includegraphics[scale=0.5]{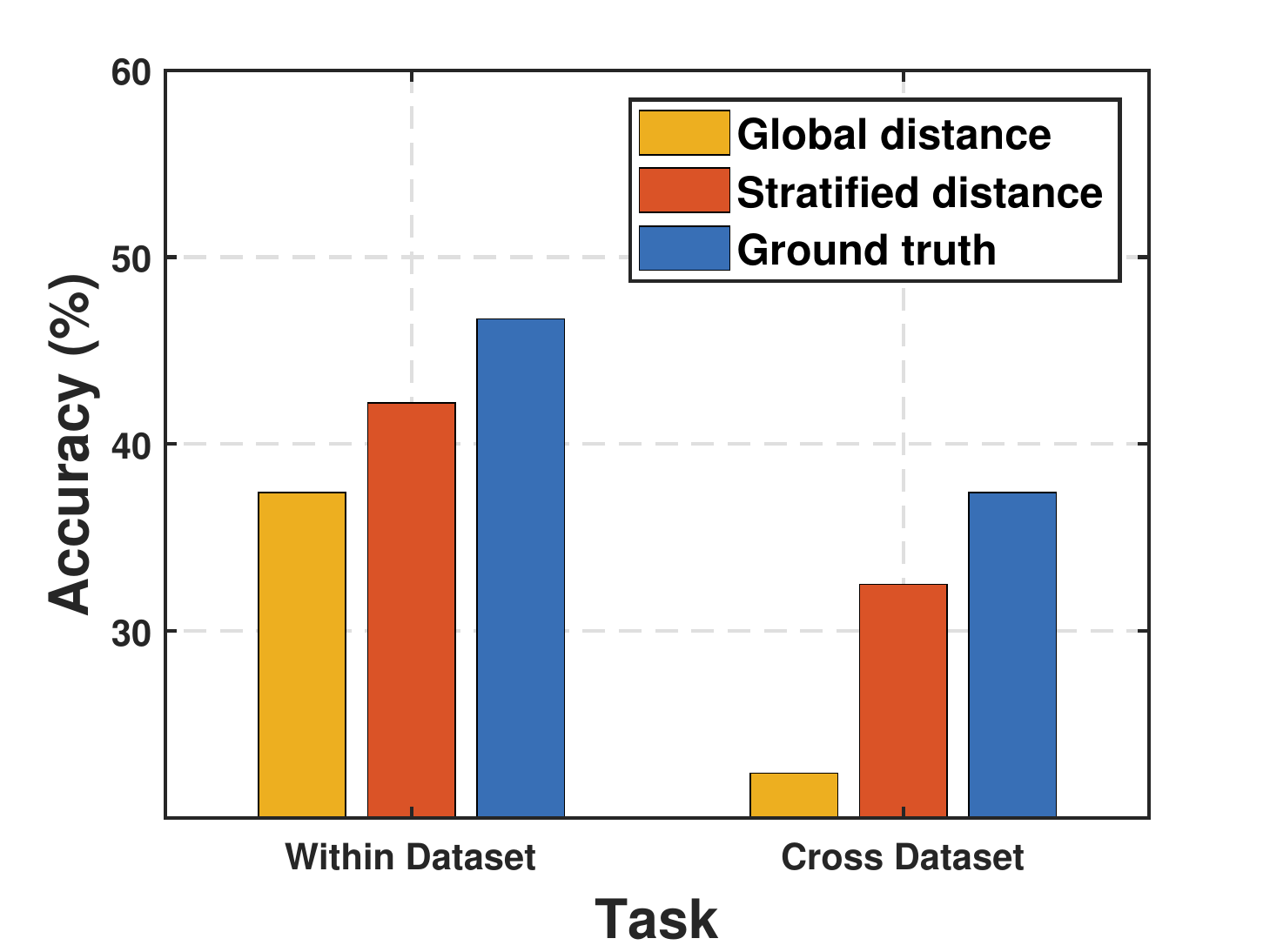}
	\caption{The comparison between global and stratified distance as the similarity measurement for domain selection.}
	\label{fig-sds}
\end{figure}

Figure~\ref{fig-sds} shows the average source domain selection accuracy of the global and stratified distance. It is obvious that our proposed SD distance achieves better performance than the traditional global distance. Note that in both \textit{Within Dataset} and \textit{Cross Dataset} scenarios, SD distance outperforms GD. Furthermore, for the even challenging \textit{Cross Dataset} distance where the similarity between the source and the target domains are little, SD could significantly outperform the traditional GD distance. This indicates the effectiveness of our proposed STL-SDS algorithm.

\begin{table}[t!]
	\centering
	\caption{Comparison of Global (G) and Stratified (S) distance on source domain selection. The symbol \boxed{item} denotes that the item is \textit{not} consistent with the ground truth.}
	\label{tb-uds}
	\resizebox{1\textwidth}{!}{
		\begin{tabular}{c|c|c|c|c|c|c|c|c|c|c|c|c|}
			\cline{2-13}
			& \multicolumn{2}{c|}{Target Dataset} & \multicolumn{5}{c|}{DSADS} & \multicolumn{5}{c|}{OPP} \\ \cline{2-13} 
			& \multicolumn{2}{c|}{Target Position} & T & RA & LA & RL & LL & B & RUA & RLA & LUA & LLA \\ \cline{2-13} 
			& \multirow{3}{*}{\begin{tabular}[c]{@{}c@{}}Cross\\ Dataset\end{tabular}} & \textbf{Truth} & RA & LA & RA & LL & RL & LUA & RLA & RUA & B & LUA \\ \cline{3-13} 
			&  & \textbf{GD} & \boxed{LA} & \boxed{T} &  \boxed{T} & LL & RL & LUA & \boxed{LLA} & RUA & \boxed{LLA} & \boxed{RUA} \\ \cline{3-13} 
			&  & \textbf{SD} & \boxed{LA} & LA & \boxed{T} & LL & RL & LUA & RLA & RUA & \boxed{LLA} & \boxed{RUA} \\ \cline{2-13} 
			& \multirow{3}{*}{\begin{tabular}[c]{@{}c@{}}Cross\\ Dataset\end{tabular}} & \textbf{Truth} & Chest & Ankle & Chest & Hand & Hand & Chest & Chest & Chest & Chest & Chest \\ \cline{3-13} 
			&  & \textbf{GD} & Chest & \boxed{Chest} & Chest & \boxed{Chest} & \boxed{Chest} & Chest & \boxed{Ankle} & \boxed{Ankle} & \boxed{Ankle} & \boxed{Ankle} \\ \cline{3-13} 
			&  & \textbf{SD} & Chest & Ankle & Chest & Hand & \boxed{Chest} & Chest & Chest & \boxed{Ankle} & Chest & Chest \\ \cline{2-13} 
		\end{tabular}
	}
\end{table}

\begin{table}[t!]
	\centering
	\caption{Accuracy (\%) of Global (GD) and Stratified (SD) distance on source domain selection.}
	\label{tb-sds-acc}
	\resizebox{1\textwidth}{!}{
	\begin{tabular}{|c|c|c|c|c|c|c|c|c|c|c|c|c|}
		\hline
		\multicolumn{2}{|c|}{Target Dataset} & \multicolumn{5}{c|}{DSADS} & \multicolumn{5}{c|}{OPP} & \textbf{Average} \\ \hline
		\multicolumn{2}{|c|}{Target Position} & T & RA & LA & RL & LL & B & RUA & RLA & LUA & LLA & - \\ \hline
		\multicolumn{2}{|c|}{Target as source} & 87.6 &	83.4 &	84.8 &	87.9 &	88.4 &	37 &	37.3 &	36.7 &	34.8 &	38.3 & 61.6 \\ \hline \hline
		\multirow{3}{*}{\begin{tabular}[c]{@{}c@{}}Within\\ Dataset\end{tabular}} & \textbf{Truth} & 35.0 & 69.5 & 66.9 & 69.9 & 77.3 & 30.9 & 33.2 & 32.1 & 24.6 & 27.8 & \textbf{46.7} \\ \cline{2-13} 
		& \textbf{GD} & \boxed{31.5} & \boxed{34.1} & \boxed{36.0} & 69.9 & 77.3 & 30.9 & \boxed{20.3} & 32.1 & \boxed{22.7} & \boxed{19.6} & 37.4 \\ \cline{2-13} 
		& \textbf{SD} & \boxed{31.5} & 69.5 & \boxed{36.0} & 69.9 & 77.3 & 30.9 & 33.2 & 32.1 & \boxed{22.7} & \boxed{19.6} & \textbf{42.2} \\ \hline \hline
		\multirow{3}{*}{\begin{tabular}[c]{@{}c@{}}Cross\\ Dataset\end{tabular}} & \textbf{Truth} & 23.8 & 25.0 & 24.8 & 25.0 & 25.0 & 50.1 & 50.0 & 49.9 & 50.0 & 50.0 & \textbf{37.4} \\ \cline{2-13} 
		& \textbf{GD} & 23.8 & \boxed{0.6} & \boxed{24.8} & \boxed{2.7} & \boxed{2.0} & 50.1 & \boxed{24.5} & \boxed{24.5} & \boxed{47.3} & \boxed{23.6} & 22.4 \\ \cline{2-13} 
		& \textbf{SD} & 23.8 & 25.0 & 24.8 & 25.0 & \boxed{2.0} & 50.1 & 50.0 & \boxed{24.5} & 50.0 & 50.0 & \textbf{32.5} \\ \hline
	\end{tabular}
}
\end{table}

Now we dig deeper into the results. Table~\ref{tb-uds} shows the selected source domains of GD and SD, and their accuracies are in Table~\ref{tb-sds-acc}. We also report the accuracy when target itself is as the source domain for comparison (which is the ideal state that can never be satisfied). From these tables, we observe: 1) The proposed Stratified distance can select better source domain than the traditional global distance with close performance with the ground truth. 2) The situation when the target domain as the source domain can achieve the best performance. However, it is only the ideal state since there are always the label scarcity problems. It indicates that CPAR is a challenging task since both of the \textit{Within Dataset} and \textit{Cross Dataset} aspects only achieve worse performance, which ensures the necessity of transfer learning algorithm. 2) Generally speaking, the most similar body parts to a side Arm or Leg is its other side. This observation is much easier to understand as common sense. 3) Torso (or Back / Chest) is the most similar body part to other body parts such as Arms and Legs. This is probably because the Torso is physically connected to Arms and Legs, leading to similar moving patterns. Therefore, most of the target domains can leverage the labeled information from Torso to build models. 4) It is important to notice that although SD achieves good results, its performance is not 100\% right. This indicates that it is extremely difficult to perform source selection. We expect to increase the performance of SD in future research.

\subsection{Evaluation of Stratified Activity Transfer}
We evaluate the performance of STL-SAT in both \textit{Within Dataset} and \textit{Cross Dataset} aspects. The state-of-the-art comparison methods include:

\begin{itemize}[noitemsep,nolistsep]
	\item PCA: Principal component analysis~\cite{fodor2002survey}.
	\item KPCA: Kernel principal component analysis~\cite{fodor2002survey}.
	\item TCA: Transfer component analysis~\cite{pan2011domain}.
	\item GFK: Geodesic flow kernel~\cite{gong2012geodesic}.
	\item TKL: Transfer kernel learning~\cite{long2015domain}.
\end{itemize}

PCA and KPCA are classic dimensionality reduction methods, while TCA, GFK, and TKL are representative transfer learning approaches. The codes of PCA and KPCA are provided in Matlab. The codes of TCA, GFK, and TKL can be obtained online \footnote{\url{https://tinyurl.com/y79j6twy}}.

We construct several CPAR tasks according to each scenario and use the labels for the target domain only for testing by following the common setting in existing work~\cite{cook2013transfer,zhao2011cross}. Other than TKL, all other methods require dimensionality reduction. Therefore, they were tested using the same dimension. After that, a classifier with the same parameter is learned using the source domain and then the target domain can be labeled. To be more specific, we use the random forest classifier~($\#Tree=30$) as the final classifier for all the 6 methods. For majority voting in STL-SAT, we simply use SVM~($C=100$), kNN~($k=3$), and random forest~($\#Tree=30$) as the base classifiers. Other parameters are searched to achieve their optimal performance. The \#iteration is set to be $T=10$ for SAT. Other parameters of STL-SAT are set $\lambda=1.0, d=30, kernel=linear$. It is noticeable that we randomly shuffle the experimental data 5 times in order to gain robust results.

Classification $accuracy$ on the target domain is adopted as the evaluation metric, which is widely used in existing transfer learning methods~\cite{long2015domain,pan2011domain}
\begin{equation}
Accuracy = \frac{|\mathbf{x}: \mathbf{x} \in \mathcal{D}_t \wedge \hat{y}(\mathbf{x})=y(\mathbf{x})|}{|\mathbf{x}:\mathbf{x} \in \mathcal{D}_t|}
\end{equation}
where $y(\mathbf{x})$ and $\hat{y}(\mathbf{x})$ are the truth and predicted labels, respectively.

Additionally, we also use the F1 score as another measurement of the results. F1 score can be calculated as
\begin{equation}
	F1=\frac{2PR}{P + R}
\end{equation}
where $P, R$ are the precision and recall, respectively.

We run STL-SAT and other methods on all tasks and report the classification accuracy in TABLE~\ref{tb-sat}. The F1 scores of all the methods are in Table~\ref{tb-f1}. For brevity, we only report the results of $A \rightarrow B$  since its result is close to $B \rightarrow A$. It is obvious that SAT significantly outperforms other methods in most cases~(with a remarkable improvement of $\mathbf{7.7}\%$ over the best baseline GFK). Compared to traditional dimensionality reduction methods~(PCA and KPCA), STL-SAT improves the accuracy by $10\% \sim 20\%$, which implies that SAT is better than typical dimensionality reduction methods. Compared to transfer learning methods~(TCA, GFK, and TKL), STL-SAT still shows an improvement of $5\% \sim 15\%$. Therefore, STL-SAT is more effective than all the comparison methods in most cases.

\begin{table}[t!]
	\centering
	\caption{Classification accuracy (\%) of STL-SAT and other comparison methods.}
	\label{tb-sat}
	\resizebox{1.0\textwidth}{!}{
		\begin{tabular}{|c|c|c|c|c|c|c|c|c|}
			\hline
			\multicolumn{2}{|c|}{Dataset} & Task & PCA & KPCA & TCA & GFK & TKL & STL-SAT \\ \hline \hline
			\multirow{8}{*}{\textit{Within Dataset}} & \multirow{3}{*}{DSADS} & RA $\rightarrow$ LA & 59.9 & 62.2 & 66.2 & 71.0 & 54.1 & \textbf{71.1} \\ \cline{3-9} 
			&  & RL $\rightarrow$ LL & 69.5 & 70.9 & 75.1 & 79.7 & 61.6 & \textbf{81.6} \\ \cline{3-9} 
			&  & RA $\rightarrow$ T & 38.9 & 30.2 & 39.4 & 44.2 & 32.7 & \textbf{45.6} \\ \cline{2-9} 
			& \multirow{4}{*}{OPP} & RUA $\rightarrow$ LUA & 76.1 & 65.6 & 76.9 & 74.6 & 66.8 & \textbf{84.0} \\ \cline{3-9} 
			&  & RLA $\rightarrow$ LLA & 62.2 & 66.5 & 60.6 & 74.6 & 66.8 & \textbf{83.9} \\ \cline{3-9} 
			&  & RLA $\rightarrow$ T & \textbf{59.1} & 47.0 & 55.4 & 48.9 & 47.7 & 56.9 \\ \cline{3-9} 
			&  & RUA $\rightarrow$ T & 68.0 & 54.5 & 67.5 & 66.1 & 60.5 & \textbf{75.2} \\ \cline{2-9} 
			& PAMAP & H $\rightarrow$ C & 35.0 & 24.4 & 34.9 & 36.2 & 35.7 & \textbf{43.5} \\ \hline
			\multirow{3}{*}{\textit{Cross Dataset}} & PAMAP $\rightarrow$ OPP & C $\rightarrow$ B & 32.8 & 43.8 & 39.0 & 27.6 & 35.6 & \textbf{40.1} \\ \cline{2-9} 
			& DSADS $\rightarrow$ P & T $\rightarrow$ C & 23.2 & 18.0 & 23.7 & 19.4 & 21.7 & \textbf{37.8} \\ \cline{2-9} 
			& OPP $\rightarrow$ D & B $\rightarrow$ T & 44.3 & 49.4 & 46.9 & 48.1 & 52.8 & \textbf{55.5} \\ \hline \hline
			\multicolumn{3}{|c|}{Average} & 51.7 & 48.4 & 53.2 & 53.7 & 48.7 & \textbf{61.4} \\ \hline
		\end{tabular}
	}
\end{table}

\begin{table}[t!]
	\centering
	\caption{F1 score of STL-SAT and other comparison methods.}
	\label{tb-f1}
	\resizebox{1.0\textwidth}{!}{
		\begin{tabular}{|c|c|c|c|c|c|c|c|c|}
			\hline
			\multicolumn{2}{|c|}{Dataset} & Task & PCA & KPCA & TCA & GFK & TKL & STL-SAT \\ \hline \hline
			\multirow{8}{*}{\textit{Within Dataset}} & \multirow{3}{*}{DSADS} & RA $\rightarrow$ LA & 0.72 & 0.72 & 0.76 & 0.73 & 0.56 & \textbf{0.81} \\ \cline{3-9} 
		&  & RL $\rightarrow$ LL & 0.72 & 0.73 & 0.77 & 0.73 & 0.67 & \textbf{0.69} \\ \cline{3-9} 
		&  & RA $\rightarrow$ T & 0.62 & 0.62 & 0.56 & 0.62 & 0.58 & \textbf{0.77} \\ \cline{2-9} 
		& \multirow{4}{*}{OPP} & RUA $\rightarrow$ LUA & 0.72 & 0.73 & 0.79 & 0.73 & 0.68 & \textbf{0.80} \\ \cline{3-9} 
		&  & RLA $\rightarrow$ LLA & 0.60 & 0.61 & 0.68 & 0.62 & 0.55 & \textbf{0.70} \\ \cline{3-9} 
		&  & RLA $\rightarrow$ T & 0.47 & 0.47 & 0.56 & 0.48 & 0.49 & \textbf{0.59} \\ \cline{3-9} 
		&  & RUA $\rightarrow$ T & 0.64 & 0.67 & 0.70 & 0.66 & 0.58 & \textbf{0.75} \\ \cline{2-9} 
		& PAMAP & H $\rightarrow$ C & 0.24 & 0.25 & 0.27 & 0.25 & 0.26 & \textbf{0.33} \\ \hline
		\multirow{3}{*}{\textit{Cross Dataset}} & PAMAP $\rightarrow$ OPP & C $\rightarrow$ B & 0.32 & 0.42 & 0.37 & 0.27 & 0.27 & \textbf{0.44} \\ \cline{2-9} 
		& DSADS $\rightarrow$ P & T $\rightarrow$ C & 0.22 & 0.20 & 0.25 & 0.24 & 0.23 & \textbf{0.39} \\ \cline{2-9} 
		& OPP $\rightarrow$ D & B $\rightarrow$ T & 0.46 & 0.50 & 0.48 & 0.50 & 0.54 & \textbf{0.59} \\ \hline \hline
		\multicolumn{3}{|c|}{Average} & 0.52 & 0.54 & 0.56 & 0.53 & 0.49 & \textbf{0.60} \\ \hline
		\end{tabular}
	}
\end{table}

The performance of TKL is the worst, because of the instability of the transfer kernel. TCA only learns a global domain shift, thus the similarity within classes is not fully exploited. The performance of GFK is second to SAT, even if GFK also learns a global domain shift. Because the geodesic distance in high-dimensional space is capable of preserving the intra properties of domains. The differences between STL-SAT and GFK are: 1)~STL-SAT outperforms GFK in most cases with significant improvement; 2)~STL-SAT strongly outperforms GFK in \textit{Within Dataset} category, indicating that SAT is more robust in recognizing different levels of activities. For SAT, it performs intra-class knowledge transfer after generating pseudo labels for candidates. Thus, better performance can be achieved by exploiting the local property of classes.

\begin{figure}[htbp]
	\hspace{-.4in}
	\includegraphics[scale=0.38]{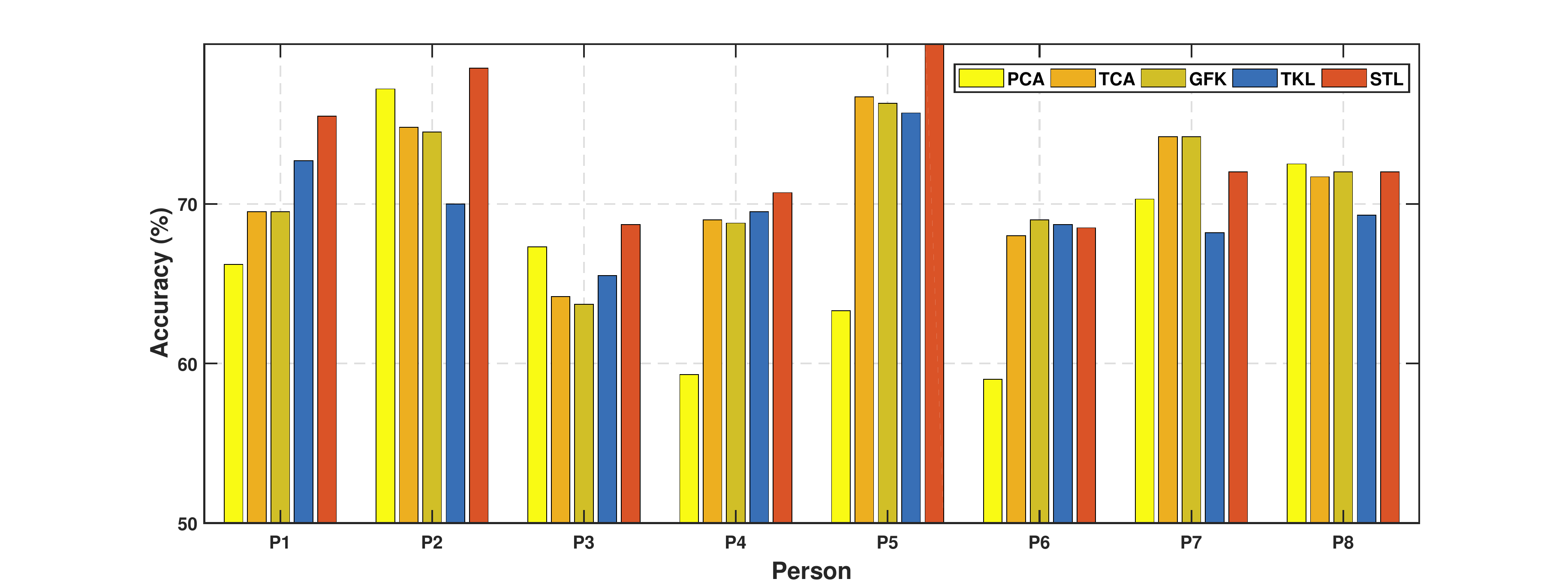}
	\vspace{-.1in}
	\caption{Classification accuracy of different methods of each person on RA $\rightarrow$ LA task.}
	\label{fig-person}
\end{figure}

Figure~\ref{fig-person} shows the classification accuracy of different methods of each person on RA $\rightarrow$ LA task of the DSADS dataset. It clearly indicates that STL-SAT achieves the best performance on most persons. On other datasets, the results are the same. We further did a case study by showing the confusion matrices of all methods for task RUA $\rightarrow$ LUA on OPP dataset in Table~\ref{tb-confusion}. The confusion matrix helps to analyze which class is easier to classify. From the results, we can observe that in CPAR, all the classes are easy to be wrongly classified since there is distribution divergence between domains. Of all the classes, \textit{Walking} and \textit{Standing} are two classes that are more easily to be misclassified. In this situation, our proposed STL-SAT could significantly outperform other comparison methods. In other situations, STL-SAT can still obtain comparable performance. Overall, STL-SAT achieves the best classification accuracy. On other tasks, the results are almost the same. It indicates the superiority of STL-SAT. We also noticed that on some persons (P6, P7, and P8), the performance of STL-SAT is worse. This may be because the sensor data of these three persons are a little mixed during the activity transition.

\begin{table}[t!]
	\centering
	\caption{Confusion matrix of STL-SAT and other comparison methods on RUA $\rightarrow$ LUA of OPP dataset. Other tasks are following the same tendency. In the table, symbols a, b, c, and d refer to four common activities: \textit{Walking, Lying, Standing,} and \textit{Sitting}}.
	\label{tb-confusion}
	\begin{tabular}{ccccccccccc}
		\cline{1-5} \cline{7-11}
		\multicolumn{1}{|c|}{PCA} & \multicolumn{4}{c|}{\textit{Acc}=72.5} & \multicolumn{1}{c|}{} & \multicolumn{1}{c|}{KPCA} & \multicolumn{4}{c|}{\textit{Acc}=74.5} \\ \cline{1-5} \cline{7-11} 
		\multicolumn{1}{|c|}{} & \multicolumn{1}{c|}{a} & \multicolumn{1}{c|}{b} & \multicolumn{1}{c|}{c} & \multicolumn{1}{c|}{d} & \multicolumn{1}{c|}{} & \multicolumn{1}{c|}{} & \multicolumn{1}{c|}{a} & \multicolumn{1}{c|}{b} & \multicolumn{1}{c|}{c} & \multicolumn{1}{c|}{d} \\ \cline{1-5} \cline{7-11} 
		\multicolumn{1}{|c|}{a} & \multicolumn{1}{c|}{\textbf{1426}} & \multicolumn{1}{c|}{144} & \multicolumn{1}{c|}{445} & \multicolumn{1}{c|}{18} & \multicolumn{1}{c|}{} & \multicolumn{1}{c|}{a} & \multicolumn{1}{c|}{\textbf{1456}} & \multicolumn{1}{c|}{134} & \multicolumn{1}{c|}{435} & \multicolumn{1}{c|}{8} \\ \cline{1-5} \cline{7-11} 
		\multicolumn{1}{|c|}{b} & \multicolumn{1}{c|}{57} & \multicolumn{1}{c|}{\textbf{492}} & \multicolumn{1}{c|}{6} & \multicolumn{1}{c|}{0} & \multicolumn{1}{c|}{} & \multicolumn{1}{c|}{b} & \multicolumn{1}{c|}{57} & \multicolumn{1}{c|}{\textbf{492}} & \multicolumn{1}{c|}{6} & \multicolumn{1}{c|}{0} \\ \cline{1-5} \cline{7-11} 
		\multicolumn{1}{|c|}{c} & \multicolumn{1}{c|}{278} & \multicolumn{1}{c|}{10} & \multicolumn{1}{c|}{\textbf{1013}} & \multicolumn{1}{c|}{\textbf{34}} & \multicolumn{1}{c|}{} & \multicolumn{1}{c|}{c} & \multicolumn{1}{c|}{278} & \multicolumn{1}{c|}{10} & \multicolumn{1}{c|}{\textbf{1013}} & \multicolumn{1}{c|}{34} \\ \cline{1-5} \cline{7-11} 
		\multicolumn{1}{|c|}{d} & \multicolumn{1}{c|}{38} & \multicolumn{1}{c|}{1} & \multicolumn{1}{c|}{54} & \multicolumn{1}{c|}{\textbf{121}} & \multicolumn{1}{c|}{} & \multicolumn{1}{c|}{d} & \multicolumn{1}{c|}{38} & \multicolumn{1}{c|}{1} & \multicolumn{1}{c|}{54} & \multicolumn{1}{c|}{\textbf{121}} \\ \cline{1-5} \cline{7-11} 
		&  &  &  &  &  &  &  &  &  &  \\ \cline{1-5} \cline{7-11} 
		\multicolumn{1}{|c|}{TCA} & \multicolumn{4}{c|}{\textit{Acc}=76.2} & \multicolumn{1}{c|}{} & \multicolumn{1}{c|}{GFK} & \multicolumn{4}{c|}{\textit{Acc}=74.3} \\ \cline{1-5} \cline{7-11} 
		\multicolumn{1}{|c|}{} & \multicolumn{1}{c|}{a} & \multicolumn{1}{c|}{b} & \multicolumn{1}{c|}{c} & \multicolumn{1}{c|}{d} & \multicolumn{1}{c|}{} & \multicolumn{1}{c|}{} & \multicolumn{1}{c|}{a} & \multicolumn{1}{c|}{b} & \multicolumn{1}{c|}{c} & \multicolumn{1}{c|}{d} \\ \cline{1-5} \cline{7-11} 
		\multicolumn{1}{|c|}{a} & \multicolumn{1}{c|}{\textbf{1423}} & \multicolumn{1}{c|}{155} & \multicolumn{1}{c|}{445} & \multicolumn{1}{c|}{10} & \multicolumn{1}{c|}{} & \multicolumn{1}{c|}{a} & \multicolumn{1}{c|}{\textbf{1449}} & \multicolumn{1}{c|}{141} & \multicolumn{1}{c|}{435} & \multicolumn{1}{c|}{8} \\ \cline{1-5} \cline{7-11} 
		\multicolumn{1}{|c|}{b} & \multicolumn{1}{c|}{152} & \multicolumn{1}{c|}{\textbf{403}} & \multicolumn{1}{c|}{0} & \multicolumn{1}{c|}{0} & \multicolumn{1}{c|}{} & \multicolumn{1}{c|}{b} & \multicolumn{1}{c|}{57} & \multicolumn{1}{c|}{\textbf{492}} & \multicolumn{1}{c|}{6} & \multicolumn{1}{c|}{0} \\ \cline{1-5} \cline{7-11} 
		\multicolumn{1}{|c|}{c} & \multicolumn{1}{c|}{369} & \multicolumn{1}{c|}{4} & \multicolumn{1}{c|}{\textbf{936}} & \multicolumn{1}{c|}{26} & \multicolumn{1}{c|}{} & \multicolumn{1}{c|}{c} & \multicolumn{1}{c|}{272} & \multicolumn{1}{c|}{12} & \multicolumn{1}{c|}{\textbf{1015}} & \multicolumn{1}{c|}{36} \\ \cline{1-5} \cline{7-11} 
		\multicolumn{1}{|c|}{d} & \multicolumn{1}{c|}{65} & \multicolumn{1}{c|}{0} & \multicolumn{1}{c|}{30} & \multicolumn{1}{c|}{\textbf{119}} & \multicolumn{1}{c|}{} & \multicolumn{1}{c|}{d} & \multicolumn{1}{c|}{37} & \multicolumn{1}{c|}{1} & \multicolumn{1}{c|}{57} & \multicolumn{1}{c|}{\textbf{119}} \\ \cline{1-5} \cline{7-11} 
		&  &  &  &  &  &  &  &  &  &  \\ \cline{1-5} \cline{7-11} 
		\multicolumn{1}{|c|}{TKL} & \multicolumn{4}{c|}{\textit{Acc}=67.8} & \multicolumn{1}{c|}{} & \multicolumn{1}{c|}{STL} & \multicolumn{4}{c|}{\textit{Acc}=78.6} \\ \cline{1-5} \cline{7-11} 
		\multicolumn{1}{|c|}{} & \multicolumn{1}{c|}{a} & \multicolumn{1}{c|}{b} & \multicolumn{1}{c|}{c} & \multicolumn{1}{c|}{d} & \multicolumn{1}{c|}{} & \multicolumn{1}{c|}{} & \multicolumn{1}{c|}{a} & \multicolumn{1}{c|}{b} & \multicolumn{1}{c|}{c} & \multicolumn{1}{c|}{d} \\ \cline{1-5} \cline{7-11} 
		\multicolumn{1}{|c|}{a} & \multicolumn{1}{c|}{\textbf{1423}} & \multicolumn{1}{c|}{157} & \multicolumn{1}{c|}{443} & \multicolumn{1}{c|}{10} & \multicolumn{1}{c|}{} & \multicolumn{1}{c|}{a} & \multicolumn{1}{c|}{\textbf{1658}} & \multicolumn{1}{c|}{67} & \multicolumn{1}{c|}{303} & \multicolumn{1}{c|}{5} \\ \cline{1-5} \cline{7-11} 
		\multicolumn{1}{|c|}{b} & \multicolumn{1}{c|}{114} & \multicolumn{1}{c|}{\textbf{430}} & \multicolumn{1}{c|}{11} & \multicolumn{1}{c|}{0} & \multicolumn{1}{c|}{} & \multicolumn{1}{c|}{b} & \multicolumn{1}{c|}{53} & \multicolumn{1}{c|}{\textbf{496}} & \multicolumn{1}{c|}{6} & \multicolumn{1}{c|}{0} \\ \cline{1-5} \cline{7-11} 
		\multicolumn{1}{|c|}{c} & \multicolumn{1}{c|}{397} & \multicolumn{1}{c|}{26} & \multicolumn{1}{c|}{\textbf{830}} & \multicolumn{1}{c|}{82} & \multicolumn{1}{c|}{} & \multicolumn{1}{c|}{c} & \multicolumn{1}{c|}{357} & \multicolumn{1}{c|}{6} & \multicolumn{1}{c|}{\textbf{912}} & \multicolumn{1}{c|}{60} \\ \cline{1-5} \cline{7-11} 
		\multicolumn{1}{|c|}{d} & \multicolumn{1}{c|}{50} & \multicolumn{1}{c|}{1} & \multicolumn{1}{c|}{40} & \multicolumn{1}{c|}{\textbf{123}} & \multicolumn{1}{c|}{} & \multicolumn{1}{c|}{c} & \multicolumn{1}{c|}{33} & \multicolumn{1}{c|}{0} & \multicolumn{1}{c|}{58} & \multicolumn{1}{c|}{\textbf{123}} \\ \cline{1-5} \cline{7-11} 
	\end{tabular}
\end{table}

\subsection{Further Analysis}
In this section, we conduct further experiments to discover more insights in cross-position activity recognition.

\subsubsection{Performance on Different Degrees of Similarities}
\label{sec-exp-sim}
\begin{figure*}[t!]
	\centering
	\subfigure[Different degrees of similarities]{
		\centering
		\includegraphics[scale=0.4]{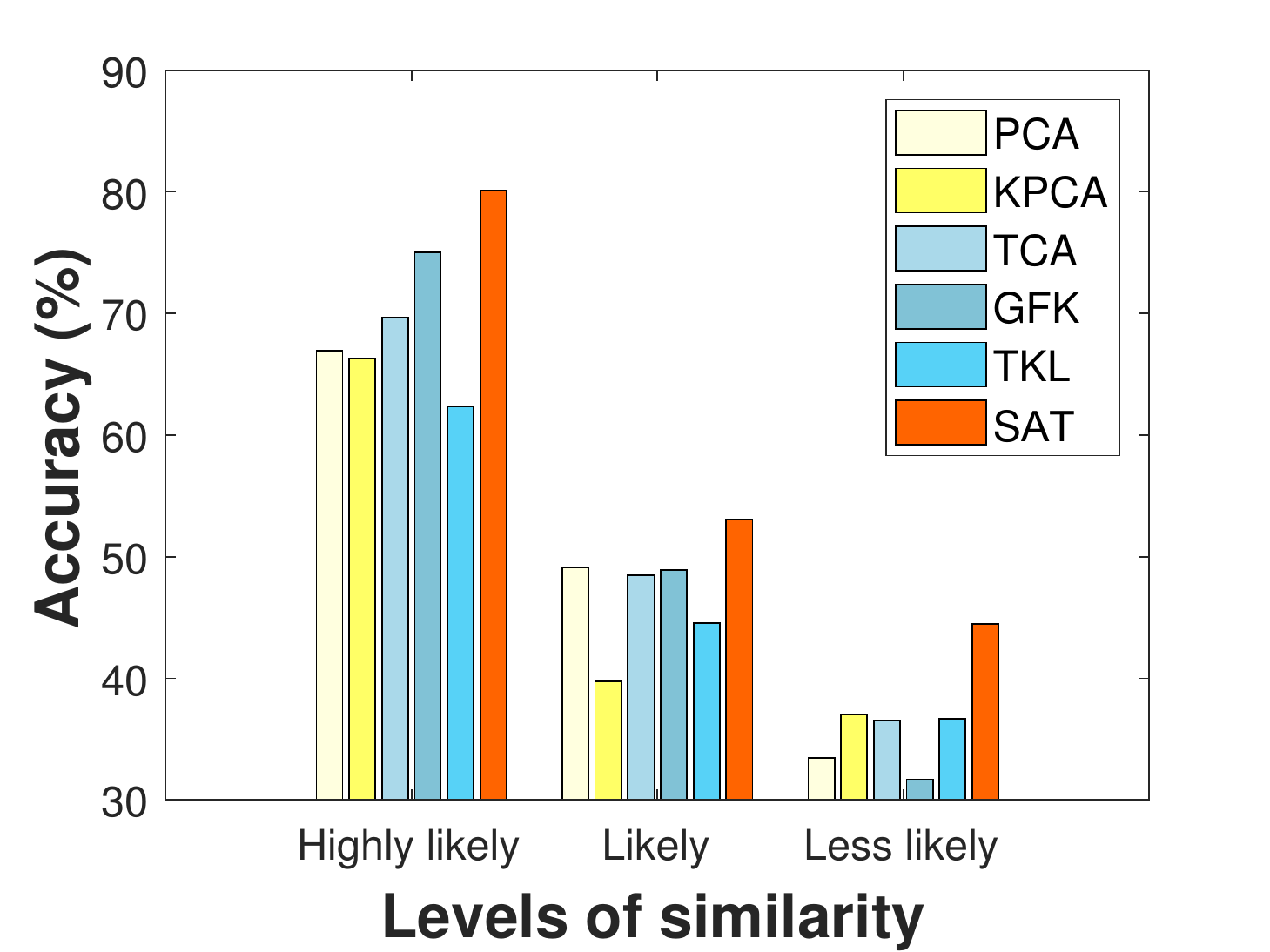}
		\label{fig-sub-sim}}
		\hspace{-.2in}
	\subfigure[Different levels of activities of 3 persons]{
		\centering
		\includegraphics[scale=0.4]{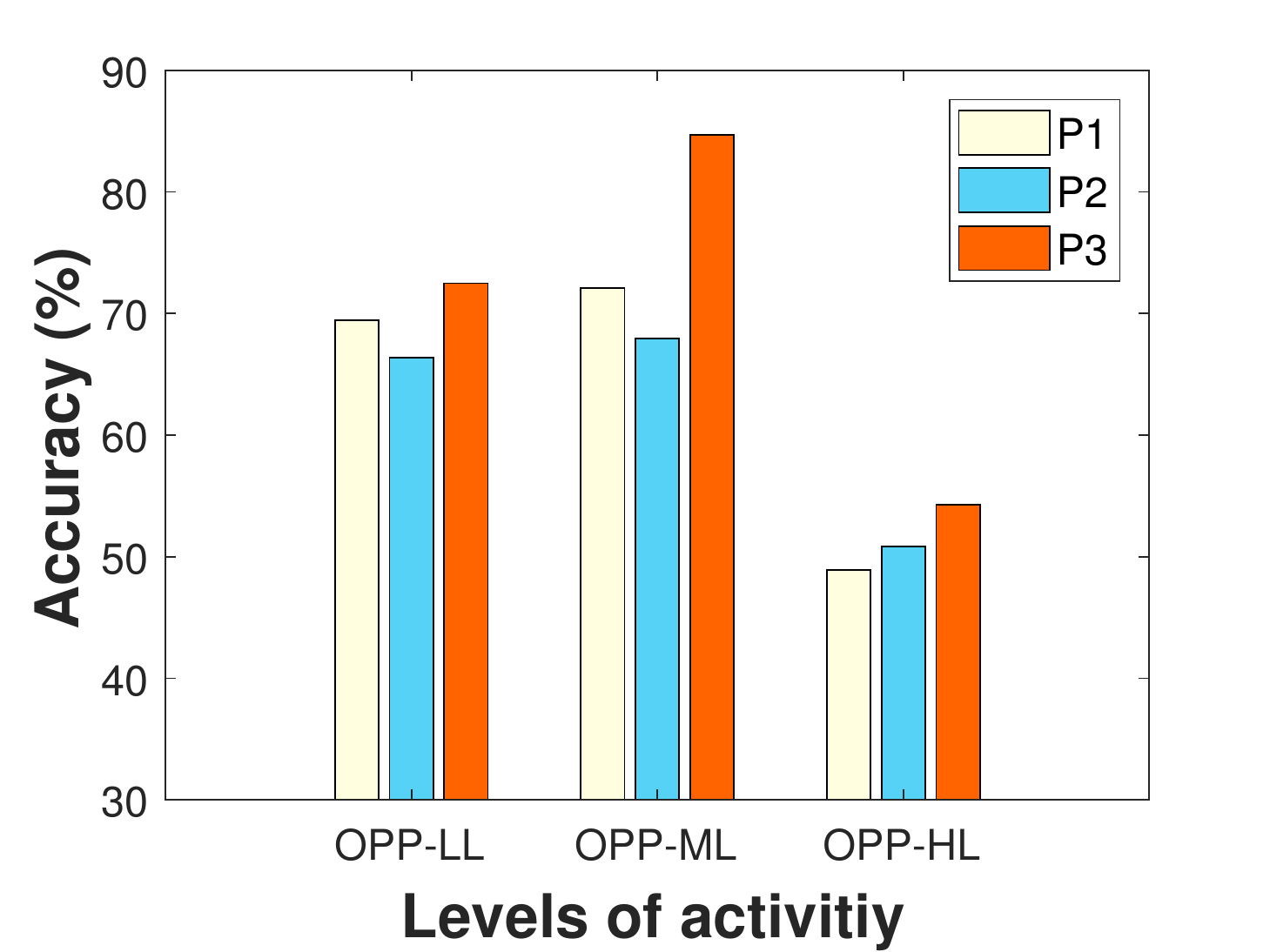}
		\label{fig-sub-level}}
	\label{fig-sim}
	\caption{Classification accuracy of~(a)~different degrees of similarities and~(b)~different levels of activities in transfer learning.}
\end{figure*}

Since the similarity between positions is extremely important to CPAR, we want to explore the performance of transfer learning methods in more fine-grained similarity degrees. Generally speaking, positions in \textit{Within Dataset} are with more similarity than \textit{Cross Dataset}. But in one dataset, can we find more degrees of similarities? It seems that in one person, Right Arm is more similar to Left Arm than to Torso. In order to explore this, we average the performance of all the methods in Figure~\ref{fig-sub-sim}. We use \textit{Highly Likely} to denote the similar body parts such as RA $\rightarrow$ LA, \textit{Likely} to denote the dissimilar body parts such as RA $\rightarrow$ T, and \textit{Less Likely} to denote the \textit{Within Dataset} for notational consistency.

For all the methods, the accuracy drops as the domain similarity becomes less. Additionally, the performance of STL-SAT is the best in all scenarios. In different degrees of similarities, the change of classification accuracy of each method follows the same tendency. Specifically, the performance of all the methods is the best between similar body parts for the same person~(e.g. RA $\rightarrow$ LA). The performance becomes worse for different body parts~(e.g. RA $\rightarrow$ T). This is because Right Arm~(RA) is more similar to Left Arm~(LA) than to Torso~(T). For a different person, all methods produce the worst results because different people have exactly different body structure and moving patterns~(e.g.~T $\rightarrow$ T across datasets). At the same degree of similarity, the results are also different. For example, the performance of RUA $\rightarrow$ T is better than RLA $\rightarrow$ T in the same dataset. Because there is more similarity between Right Upper Arm~(RUA) and Torso~(T) than between Right Lower Arm~(RLA) and Torso~(T). Other positions also share a similar discovery.

All the experimental results indicate that the \textit{similarity} between the source and target domain is important for cross-domain learning. In real life, other factors such as age and hobby also help define the similarity of activities. For other cross-domain tasks~(image classification etc.), finding the relevant domain is also important. For example, the most similar image set for a dog is probably a cat, since those two kinds of animals have similar body structures, moving patterns, and living environments.

\subsubsection{Performance on Different Levels of Activities}
\label{sec-exp-activity}
The previous experiments are mainly on the activity of daily living, which is mostly body movements and sports activities. In real life, there are more than this level of activities. For instance, \textit{Having Coffee} is much higher than \textit{Walking} since it requires more environmental information. Different levels of activities imply the different depths of activity granularities. In this section, we extensively investigate the performance of CPAR on different levels of activities. By taking advantage of the diverse activity classes in OPP dataset~\cite{chavarriaga2013opportunity}, we analyze the transfer learning performance on low-level~(\textbf{OPP-LL}), middle-level~(\textbf{OPP-ML}), and high-level~(\textbf{OPP-HL}) activities. The results are presented in Figure~\ref{fig-sub-level}. It indicates the best performance is achieved at \textit{middle-level} activities, while it suffers from low-level and even worse at high-level activities.

Low-level activities such as \textit{Walking} and middle-level activities such as \textit{Closing} are mostly contributed by the atomic movements of body parts and they are likely to achieve better transfer results than the high-levels. On the other hand, high-level activities such as \textit{Coffee Time} not only involve basic body movements but also contain \textit{contextual} information like ambient or objects, which is difficult to capture only by the body parts. Since the bridge of successful cross-position transfer learning is the similarity of body parts, it is not ideal to achieve good transfer performance by only using body parts. The reason why results on OPP-ML are better than OPP-LL is that activities of OPP-ML are more fine-grained than OPP-LL, making it more capable of capturing the similarities between the body parts.

\subsection{Effectiveness Analysis}
In this section, we verify the effectiveness of STL-SAT in several aspects since it is more complicated than STL-SDS. The core idea of STL-SAT is \textit{intra-class transfer}, where the \textit{pseudo} labels of the \textit{candidates} are acting as the evidence of transfer learning. It is intuitive to ask the following three questions. Firstly, \textit{Can the confidence of the pseudo labels have an influence on the algorithm?} It seems that STL-SAT will not achieve good performance if there is not enough \textit{reliable} candidates available. Secondly, \textit{Can the choice of majority voting classifiers affect the framework?} If we use different classifiers, the performance is likely to vary. Thirdly, \textit{Is the performance of STL-SAT robust to parameter selection?} There are several parameters in STL-SAT: trade-off parameter $\lambda$ and dimension $m$. Will they affect the performance? In this part, we answer these questions through the following experiments. 

\begin{figure*}[t!]
	\centering
	\subfigure[Candidate usage]{
		\centering
		\includegraphics[scale=0.26]{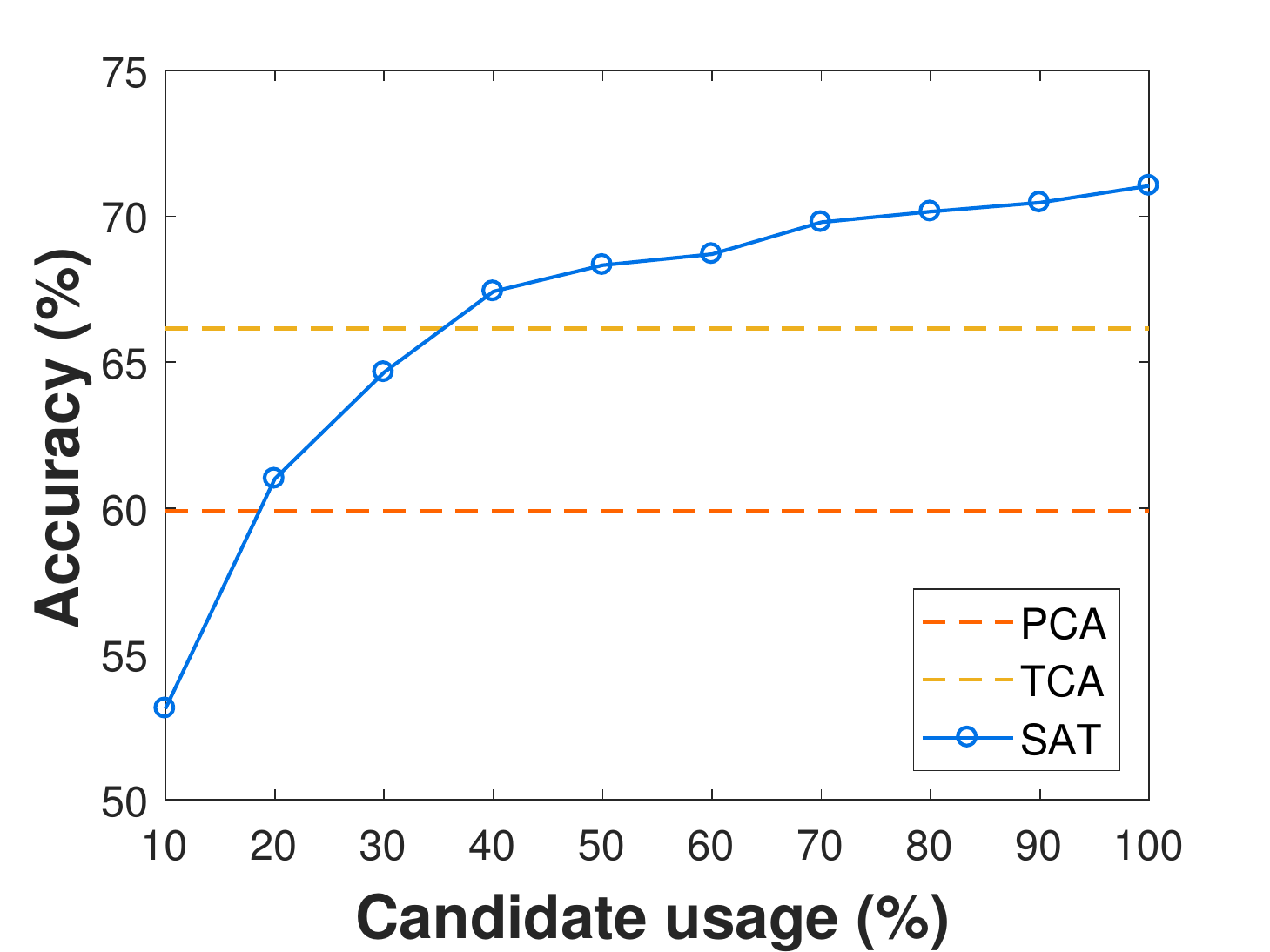}
		\label{fig-sub-candidate}}
	\hspace{-.2in}
	\subfigure[Iteration]{
		\centering
		\includegraphics[scale=0.26]{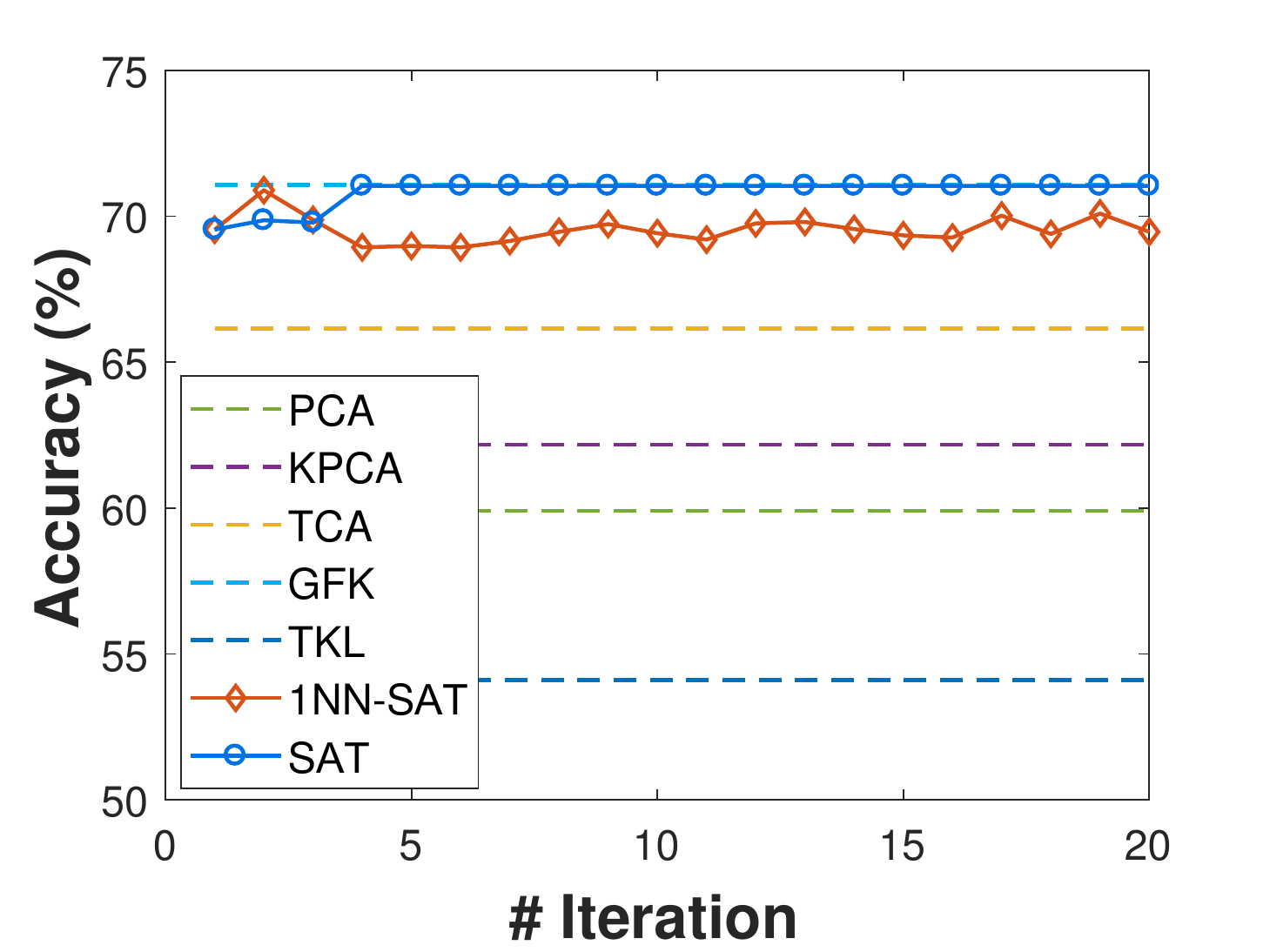}
		\label{fig-sub-ite}}
	\hspace{-.2in}
	\subfigure[Parameter sensitivity]{
		\centering
		\includegraphics[scale=0.26]{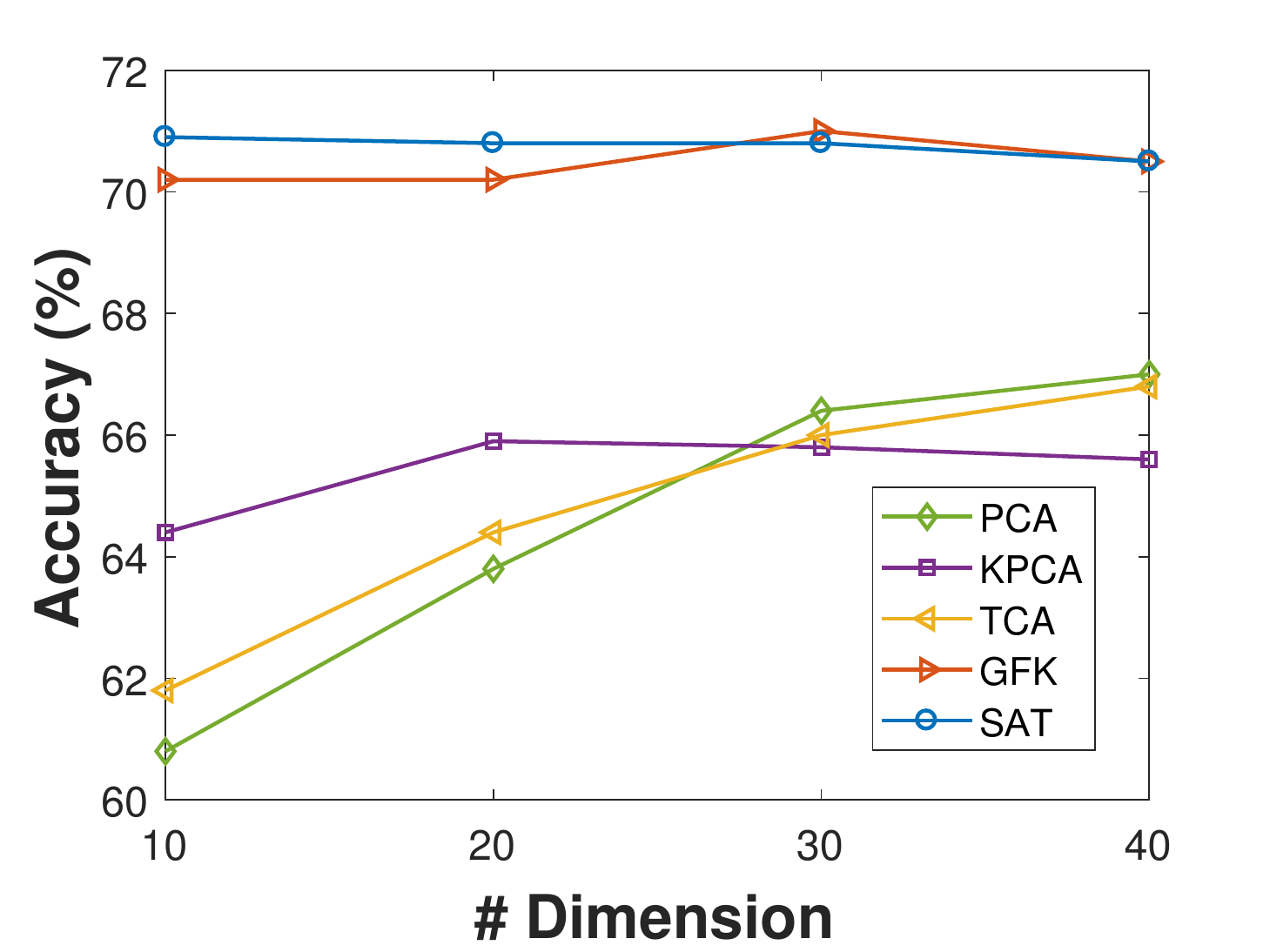}
		\label{fig-sub-para}}
	\label{fig-last}
	\caption{Detailed results: a) Candidates usage of STL-SAT for RA $\rightarrow$ LA on DSADS dataset; b) Convergence of 1NN-SAT and STL-SAT; c) Parameter sensitivity}
\end{figure*}

\textbf{1) The confidence of the candidates:} We control the percentage of the \textit{candidates} from 10\% to 100\% in every trial and make the rest belong to the \textit{residual} part. Then we test the performance of STL-SAT. We test on the task LA $\rightarrow$ RA on DSADS and compare with other 2 methods (PCA and TCA). The result is shown in Figure~\ref{fig-sub-candidate}. For simplicity, we only compare STL-SAT with PCA and TCA, since they are both classic dimensionality reduction methods. From the results, we can observe that the performance of STL-SAT is increasing along with the increment of candidates percentage. More importantly, STL-SAT outperforms the other two methods with \textbf{less than} 40\% of candidates. It reveals that SAT does not largely rely on the confidence of the candidates and can achieve good performance even with fewer candidates.

On the other hand, we also noticed that the accuracy of the majority voting classifiers may significantly influence the results of STL. Luckily, a previous work~\cite{Wang2018visual} has verified that the accuracy of the voting classifiers will not heavily influence the final results. In their experiments, they chose different kinds of classifiers to generate pseudo labels. The results of different classifiers are showing that the final results will not heavily rely on the power of the classifiers.

\textbf{2) Majority voting classifiers and iteration:} To test the effectiveness of majority voting classifiers, we choose 1 nearest neighbor~(1NN) as the base classifier of majority voting. Then we run SAT iteratively. The results are shown in Figure~\ref{fig-sub-ite}, where `1NN-SAT' and `SAT' denotes the result of STL-SAT using 1NN and random forest as the base majority voting classifier, respectively. From those results, we can observe: 1) STL-SAT can iteratively improve the classification accuracy even with some weak majority voting classifier. 1NN-SAT achieves slightly worse than STLSAT, indicating that STL-SAT is rather \textit{robust} to the base classifiers. Since more powerful classifiers would lead to better performance, we strongly suggest using more reliable classifiers for majority voting in real problems. 2) STL-SAT can reach a quick convergence within \textit{fewer than} 10 iterations. This indicates that STL-SAT can be efficiently trained.

\textbf{3) Parameter sensitivity.} STL-SAT involves two parameters: the dimension $m$, and trade-off parameter $\lambda$. In this experiment, we empirically evaluate the sensitivity of $m$. The evaluation of $\lambda$ follows the same tendencies. We set $m \in \{10, 20, 30, 40\}$ and test the performance of SAT and other dimensionality reduction methods. As shown in Figure~\ref{fig-sub-para}, STL-SAT achieves the best accuracy under different dimensions. Meanwhile, the accuracy of STL-SAT almost does not change with the decrement of $m$. The results reveal that STL-SAT is much more effective and robust than other methods under different dimensions. Therefore, STL-SAT can be easily applied to many cross-domain tasks which require robust performance w.r.t. different dimensions.

\section{Potentials of STL in Pervasive Computing}
\label{sec-dis}
We studied cross-domain activity recognition through the proposed STL framework and evaluated its performance on cross-position HAR tasks. There are more applications in pervasive computing that STL could be used for. In this section, we discuss the potential of STL in other applications.

\textit{1) Activity recognition.} The results of activity recognition can be different according to different \textit{devices}, \textit{users}, and wearing \textit{positions}. STL makes it possible to perform cross-device/user/position activity recognition with high accuracy. In case cross-domain learning is needed, finding and measuring the similarity between the device/user/position is critical.

\textit{2) Localization.} In WiFi localization, the WiFi signal changes with the \textit{time}, \textit{sensor}, and \textit{environment}, causing the distributions to differ. Therefore, it is necessary to perform cross-domain localization. When applying SAT to this situation, it is also important to capture the similarity of signals according to time/sensor/environment.

\textit{3) Gesture recognition.} Due to the differences in hand structure and movement patterns, the current gesture recognition models cannot generalize well. In this case, STL can be a good option. Meanwhile, special attention needs to be paid to the divergence between the different characteristics of the subjects. 

\textit{4) Other context-aware applications.} Other applications include smart home sensing, intelligent city planning, healthcare, and human-computer interaction. They are also context-aware applications. Most of the models built for pervasive computing are only \textit{specific} to certain contexts. Transfer learning makes it possible to transfer the knowledge between related contexts, of which STL can achieve the best performance. However, when recognizing high-level contexts such as \textit{Coffee Time}, it is rather important to consider the relationship between different contexts in order to leverage their similarities. The research in this area is still on the go.

\textit{5) Suggestions for using STL.} Firstly, before using STL, it is critical to extracting useful feature representations from the original data. Secondly, it is better to use strong classifiers for majority voting to improve the convergence rate. Thirdly, the intra-class transfer step of STL-SAT can be deployed in parallel if we compute the transformation matrix of each class.

\section{Conclusions and Future Work}
\label{sec-conclu}
The label scarcity problem is very common in activity recognition. Transfer learning addresses this issue by leveraging labeled data from auxiliary domains to annotate the target domain. In this paper, we propose a novel and general Stratified Transfer Learning~(STL) framework for cross-domain learning in pervasive computing. STL can address both the source domain selection (by STL-SDS) and knowledge transfer (by STL-SAT) problems in CPAR. Compared to existing approaches which only learn a global distance, STL can exploit the local property of each class between different domains. Experiments on three large public datasets demonstrate the significant superiority of STL over five state-of-the-art methods. We extensively analyze the performance of transfer learning under different degrees of similarities and different levels of activities.

In the future, we plan to extend STL in the following research directions:

One natural idea is to extend STL in the deep neural networks, which can perform end-to-end learning for activity recognition. The biggest difference between this paper and the deep version is we can exploit deep networks to automatically extract features and then perform domain selection and activity transfer. All these steps can be operated in one neural network. In such a network, our focus will be designing effective backward strategies for STL.

Heterogeneous activity recognition. Currently, STL can be applied in a homogeneous situation where the source and the target domains are sharing the same features. In a heterogeneous situation, the features for different domains may be different. For instance, if the source domain contains activity information of the accelerometer and the target domain has information of the gyroscope, we need to develop heterogeneous activity recognition algorithms to deal with this situation. One natural idea is to use autoencoder to unitedly represent the different kinds of features. Thus, domains can share the same features at the high level of the autoencoder.

\section*{Acknowledgments}
This work is supported in part by National Key R \& D Plan of China (No.2017YFB1002802), NSFC~(No.61572471,61472399), and Beijing Municipal Science \& Technology Commission (No.Z171100000117017).

\section*{References}

{\small
	\bibliographystyle{IEEEtran}
	\bibliography{percom18}
}

\end{document}